\documentclass[10pt,twocolumn,letterpaper]{article}

\usepackage{cvpr}              
\definecolor{cvprblue}{rgb}{0.21,0.49,0.74}
\usepackage[pagebackref,breaklinks,colorlinks,allcolors=cvprblue]{hyperref}
\usepackage{colortbl}

\def\confName{CVPR}
\def\confYear{2027}

\title{LumosFlow: Latent Space Motion Generation for Video Frame Interpolation}

\author{Jiahao Chen\\
Tsinghua University\\
\and
Hangjie Yuan\\
DAMO Academy, Alibaba Group\\
\and 
Yichen Qian \\
DAMO Academy, Alibaba Group\\
\and 
Jingyun Liang \\
DAMO Academy, Alibaba Group\\
\and
Jiazheng Xing \\
Zhejiang University \\
\and
Pengwei Liu \\
Zhejiang University\\
\and 
Weihua Chen \\
DAMO Academy, Alibaba Group\\
}

\begin{document}
\maketitle
\begin{abstract}
Recent diffusion-based methods have shown remarkable performance in video frame interpolation (VFI) by first generating motion (optical flow) and then guiding frame synthesis.  However, capturing complex motion remains challenging, as learning diverse motion patterns is computationally demanding and difficult to scale. To address this issue, we propose LumosFlow, which generates complex motion in a compressed latent space, enabling high-quality and temporally coherent frame interpolation. Specifically, LumosFlow comprises two key modules: the Latent Optical Flow Diffusion Model (LOF-DM) and MotionControlNet. Trained on large-scale motion data, LOF-DM supports optical flow generation at multiple resolutions and frame rate conversion factors ($4\times$, ..., $24\times$) in latent space. MotionControlNet uses the flow generated by LOF-DM to guide the refinement of warped frames, improving the quality of synthesized frames. Experiments demonstrate that LumosFlow achieves superior performance on challenging VFI benchmarks. Building on this capability, we further apply LumosFlow to video generation, interpolating between successive keyframes to produce high-quality videos with smooth, temporally coherent motion. Code and checkpoints will be released upon acceptance.
\end{abstract}    
\section{Introduction}
Recently, with the development of diffusion model~\cite{ho2020denoising}, many researchers have adpot it for Video Frame Interpolation (VFI)~\cite{xu2019quadratic, niklaus2017video} and achieve remarkable performance. Most existing methods directly generate intermediate frames conditioned on the first and last frames, training from scratch~\cite{jain2024video, danier2024ldmvfi} or fine-tuning the Image-to-Video diffusion model~\cite{aigc_apps_VideoX_Fun_2026,peng2026towards, wang2025generative}. Although these methods can produce visually realistic frames, maintaining coherence across the interpolated sequence remains challenging, particularly under large and complex motion~\cite{ma2026bi, zhang2025eden}. Therefore, how to achieve natural and temporally coherent frame interpolation still remains an important problem.

\begin{figure}
    \centering
    \includegraphics[width=0.9\linewidth]{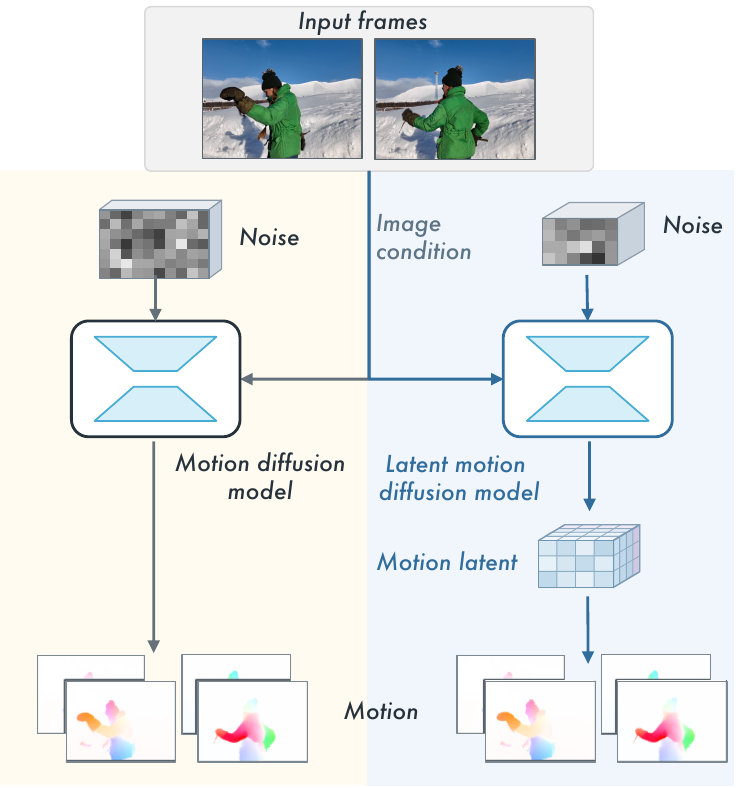}
    \vskip -0.15in
    \caption{Comparison of motion diffusion paradigms. Conditioned on the first and last frames, previous methods (left) generate optical flow directly in the original flow space, while our method (right) performs diffusion in the latent space.}
    \label{fig:1}
    \vskip -0.2in
\end{figure}

To resolve this limitation, several methods explicitly model motion to guide subsequent frame synthesis~\cite{lew2025disentangled, hai2025hierarchical, ma2026realtime}. By first generating optical flow via diffusion models,  a two-stage interpolation pipeline warps the input frames and refines the warped results to synthesize intermediate frames. For example, MoMo~\cite{lew2025disentangled} employs a motion diffusion model to generate bidirectional optical flow, which is then used by a separate frame synthesis network. Similarly, HFD~\cite{hai2025hierarchical} generates bilateral optical flow through coarse-to-fine diffusion and jointly optimizes the flow model and image synthesizer. RDVFI~\cite{ma2026realtime} uses a video diffusion model to generate sparse latent keyframes, from which continuous pixel trajectories are estimated to guide flow-based rendering. Benefiting from diffusion-based motion modeling, these methods can better capture motion patterns and use them to guide frame synthesis, improving consistency compared with direct diffusion-based frame generation~\cite{danier2024ldmvfi, jain2024video}.

\begin{figure*}
    \centering
    \includegraphics[width=0.9\linewidth]{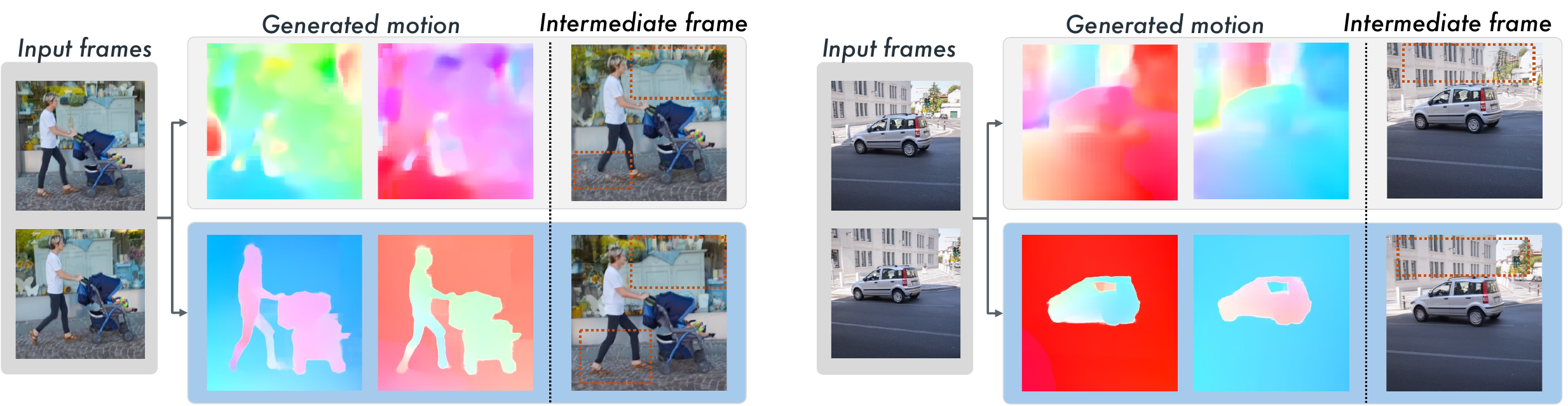}
    \vskip -0.15in
    \caption{Visual comparison between LumosFlow (highlighted in blue) and MoMo (highlighted in gray). Red boxes indicate visual failures (or artifacts) in MoMo and finer details in LumosFlow.}
    \label{fig:2}
    \vskip -0.2in
\end{figure*}

However, existing motion-guided approaches still face significant challenges in efficiency and scalability. As shown in Fig.~\ref{fig:1}, these methods typically perform diffusion directly in dense spatial domains, either on the original RGB grid or on moderately downsampled grids~\cite{hai2025hierarchical, lew2025disentangled}. Operating over such dense representations incurs substantial computational overhead at high resolutions, severely constraining training scale under a fixed budget. This limitation is particularly critical given the inherent complexity and diversity of real-world motion; insufficient training scale fails to capture the full spectrum of motion patterns, leaving rare dynamics underrepresented and hindering generalization to challenging scenes. For example, as shown in Fig.~\ref{fig:2}, MoMo~\cite{lew2025disentangled} fails to generalize well to complex dynamics, producing inaccurate optical flows on both objects and backgrounds. These estimation errors subsequently propagate through frame synthesis, resulting in distortions in the interpolated frames.

We therefore aim to develop a scalable motion diffusion foundation model that generates complex motion with large displacements. Unlike RGB videos, optical flow encodes pure pixel displacements devoid of appearance details such as color and texture, thereby exhibiting greater spatial coherence and sparser spatial variations. As discussed in Sec.~\ref{motivation}, our theoretical analysis suggests that these properties allow aggressive spatial compression while keeping reconstruction error bounded. Inspired by latent diffusion models for video generation~\cite{wan2025wan, kong2024hunyuanvideo, hacohen2024ltx}, we first introduce an Optical Flow VAE (OF-VAE) to synthesize motion within a compressed latent space. Specifically, we increase the spatial downsampling factor from the conventional $8\times$ or $16\times$ used in RGB video VAEs~\cite{yang2024cogvideox} to $32\times$, yielding a substantially more compact latent representation. This design facilitates efficient large-scale training and significantly accelerates motion generation compared to performing diffusion directly in the dense optical flow space.

Building on OF-VAE, we propose LumosFlow, a motion guided VFI framework. LumosFlow contains two key components: the Latent Optical Flow Diffusion Model (LOF-DM) and MotionControlNet. LOF-DM generates motion in the compressed latent space conditioned on the first and last frames. The decoded optical flow is then used to warp the input frames, and MotionControlNet incorporates the warped results as explicit motion guidance for frame synthesis and refinement. By combining efficient latent motion generation with motion-guided refinement, LumosFlow produces high-quality intermediate frames with natural and temporally coherent motion.

Our contributions can be summarized as follows:
\begin{itemize}
\item We compress optical flow in latent space, motivated by a theoretical analysis of motion compressibility. This compact representation enables efficient large-scale training and faster motion generation.

\item We propose LumosFlow, a motion guided VFI framework combining LOF-DM for latent optical flow generation with MotionControlNet for frame refinement, supporting multiple resolutions and frame rate conversion factors.

\item We achieve promising results on the DAVIS-7, UCF101-7, and FCVG benchmarks, with qualitative comparisons demonstrating effectiveness in handling large and complex motion. We further validate LumosFlow in keyframe-based video generation by interpolating between consecutive keyframe pairs to produce temporally coherent video sequences.

\end{itemize}
\section{Related Work}
\paragraph{Video Frame Interpolation}
Video frame interpolation (VFI) synthesizes intermediate frames for smoother motion and higher frame rates. Flow-based methods~\cite{liu2017video, bao2019depth, huang2022real} are widely used for their strong temporal consistency. Their explicit motion representations provide correspondences for aligning and combining content from the input frames. RIFE~\cite{huang2022real} directly estimates intermediate flows with a lightweight network for fast and accurate interpolation, while AMT~\cite{Li_2023_CVPR} uses all-pairs correlations and multi-scale flows for backward warping and frame fusion. Recent diffusion-based methods model intermediate flow distributions to address motion ambiguity. MoMo~\cite{lew2025disentangled} decouples motion generation from frame synthesis via bilateral flow diffusion, while HFD~\cite{hai2025hierarchical} predicts bilateral flows through a coarse-to-fine hierarchical diffusion process before synthesis. However, direct flow-space diffusion remains computationally expensive. We generate optical flows in a compact latent space, reducing computational cost while retaining explicit motion guidance for frame synthesis.

\paragraph{Video Diffusion Models}
Video diffusion models~\cite{blattmann2023stable, kong2024hunyuanvideo, wan2025wan, yang2024cogvideox} extend image diffusion to the temporal domain. They jointly model spatial appearance and temporal dynamics to synthesize coherent sequences through iterative denoising, typically in a compact latent space. Beyond text- and image-to-video generation, recent frameworks incorporate depth maps~\cite{chen2023control}, sketches~\cite{wang2023videocomposer}, reference images~\cite{chen2025humo}, and drag trajectories~\cite{yin2023dragnuwa}. These signals enable fine-grained control over the structure, appearance, and motion of generated videos. Conditioning on the first and last frames~\cite{wang2025generative} further enables transitions between specified boundary frames. However, directly synthesizing intermediate frames in latent space can produce implausible motion due to weak inter-frame correspondence constraints. We use generated optical flows to provide correspondence constraints for frame synthesis, improving motion plausibility.
\section{Motivation}
\label{motivation}
Existing video generation methods operate in a latent space obtained by compressing RGB videos along both spatial and temporal dimensions~\cite{yang2024cogvideox}. Compared with RGB video, optical flow exhibits greater spatial coherence and captures motion without explicitly encoding appearance details such as color and texture. In the following, we quantify this difference and establish conditions under which optical flow can support higher spatial compression ratios.

Let $S \in \mathbb{R}^{K\times C\times H\times W}$ denote a 3D sequence data, where $K$, $C$, $H$, and $W$ represent the number of frames, channels, height, and width, respectively. Define the spatial mean and the normalized spatial variation as
\begin{equation}
\mu_{k,c}=\frac{1}{HW}\sum_{i,j}S_{k,c,i,j}
\qquad
 \gamma(S)=\frac{\|\nabla_{xy}S\|_F^2}{\|S-\mu\|_F^2},
\end{equation}
where \(\nabla_{xy}\) contains valid horizontal and vertical adjacent differences. Then we can derive the following proposition.

\newtheorem{proposition}{Proposition}[section]
\begin{proposition}
\label{pro}
 For an integer \(h\ge2\) dividing \(H,W\), let \(\widehat S_h\) be reconstructed by averaging each non-overlapping \(h\times h\) block and replicating its mean. This construction is realizable by a strided convolutional encoder and an upsampling decoder, and satisfies
\[
\frac{\|S-\widehat S_h\|_F^2}{\|S-\mu\|_F^2}
\le\frac{h^2}{4}\gamma(S).
\]
\end{proposition}

\textit{Remark}: Prop.~\ref{pro} shows that the reconstruction error bound depends jointly on the spatial downsampling factor $h$ and the normalized spatial variation $\gamma(S)$. Increasing $h$ improves compression efficiency by reducing the spatial resolution along each dimension. However, the error bound $\frac{h^2}{4}\gamma(S)$ also increases quadratically with $h$. For sequences with higher $\gamma(S)$, a smaller $h$ is therefore needed to keep the error bound low. Conversely, sequences with lower $\gamma(S)$ can accommodate a larger $h$ while maintaining the same bound, allowing stronger spatial compression. Detailed proofs are in the Appendix Sec.A

\begin{figure}
    \centering
    \includegraphics[width=0.95\linewidth]{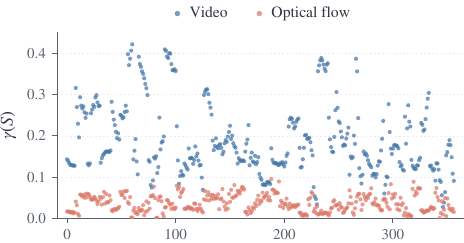}
    \vskip -0.15in
    \caption{Comparison of $\gamma(S)$ for video and optical flow on the DAVIS-7 dataset. The x-axis denotes the sample ID. }
    \label{fig:gamma}
    \vskip -0.2in
\end{figure}

We also quantify this difference on the DAVIS-7 dataset~\cite{jain2024video}. As shown in Fig.~\ref{fig:gamma}, optical flow generally exhibits lower $\gamma(S)$ values than the corresponding RGB video samples. This indicates lower normalized spatial variation, consistent with stronger spatial coherence in the optical flow fields. These results motivate us to use larger spatial downsampling factors for optical flow than for RGB video.

\begin{figure*}
    \centering
    \includegraphics[width=0.96\linewidth]{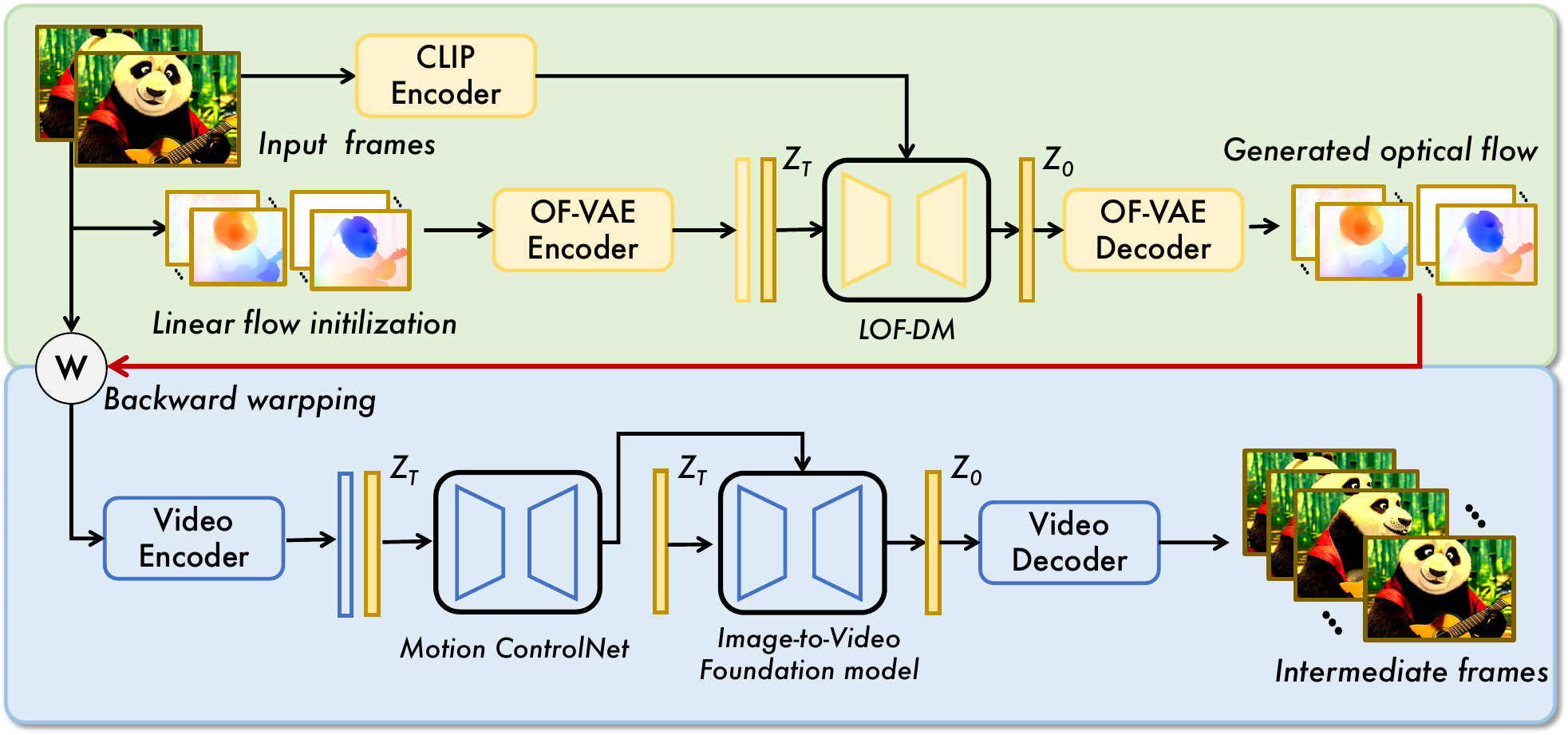}
    \vskip -0.15in
    \caption{The overall framework of LumosFlow, compriseing two components: motion generation (highlighted in green) and post-hoc refinement (highlighted in blue).}
    \label{main}
    \vskip -0.2in
\end{figure*}
\section{LumosFlow}
In this section, we present LumosFlow, a latent space motion generation method for VFI. As shown in Fig.~\ref{main}, LumosFlow is divided into two stages: optical flow generation and post-hoc refinement. In Sec.~\ref{method4.2}, we introduce the Optical Flow Variational AutoEncoder (OF-VAE), which efficiently compresses optical flow into a latent space. In Sec.~\ref{method4.3}, we describe the design of the diffusion model, which generates the optical flows. Finally, in Sec.~\ref{method4.4}, we introduce the proposed MotionControlNet for refining the warped frames to produce the final interpolated results.

\subsection{Optical Flow VAE}
\label{method4.2}
Given two images $I_1$ and $I_K$, it is difficult to synthesize intermediate frames $\{\hat{I}_k \mid k=2,\cdots, K-1\}$ at time stamp $k$ directly. Previous methods like RIFE~\citep{huang2022real} decouple intermediate frame generation into motion (optical flow) estimation and appearance refinement. Therefore, we first investigate the generation of optical flow. 
Formally, we denote $F_{k \rightarrow 1}$ and $F_{k \rightarrow K}$ as the optical flows from $I_k$ to $I_1$ and $I_K$, respectively, and $\hat{F}_{: \rightarrow 1}=\{\hat{F}_{t \rightarrow 1}\}_{t=2}^{K-1}$, $\hat{F}_{: \rightarrow K}=\{\hat{F}_{t \rightarrow K}\}_{t=2}^{K-1}$.

Optical flows and videos share the same dimensionality, the direct generation of optical flow in the pixel space still incurs substantial computational costs. Considering the information conveyed by optical flow is less rich than that of RGB data, allowing it to support greater compression ratios, we propose an Optical Flow Variational AutoEncoder (OF-VAE) that compresses optical flow within the latent space, rather than directly generating it at the pixel level. 

Formally, given a video $v \in \mathbb{R}^{K\times C \times H \times W}$, we extract the first ($I_1$) and last frames ($I_K$) as the key frames and compute the optical flow for all intermediate frames, denoted as $F_{:\rightarrow 1} \in \mathbb{R}^{(K-2) \times 2 \times H \times W}$ and $F_{:\rightarrow K}$.
Since  $F_{:\rightarrow 1}$ and $F_{:\rightarrow K}$ share common motion information, we concatenate them along the channel dimension and use an encoder $\mathcal{E}$ map them into the latent space $z = \mathcal{E}([F_{:\rightarrow 1}, F_{:\rightarrow K}])$, where $z \in \mathbb{R}^{k \times c \times h \times w}$, and a decoder $\mathcal{D}$ reconstruct optical flows, giving $[\hat{F}_{:\rightarrow 1}, \hat{F}_{:\rightarrow K}]=\mathcal{D}(z)$. Similar to Stable Diffusion~\citep{rombach2022high}, the encoder downsamples the optical flow by a factor $f = H / h = W / w$, $g = K/k$. Considering the sparsity of optical flow, we set $f=32$ and $g=4$. Compared with the previous image VAE~\citep{podell2023sdxl}, OF-VAE has a higher compression ratio. 

Finally, the optimization target is shown as follows, where $\text{KL}_{reg}$ denotes the Kullback-Leibler Divergence~\citep{kullback1951information} regularization term.
\begin{equation}
\begin{aligned}
    L_{\text{OF-VAE}} = \lVert F_{:\rightarrow 1} -   \hat{F}_{:\rightarrow 1} \rVert_1 + \lVert F_{:\rightarrow K} -   \hat{F}_{:\rightarrow K} \rVert_1 + \text{KL}_{reg}.
\end{aligned}
\end{equation}

\subsection{Latent Optical Flow Diffusion Model}
\label{method4.3}
With our trained OF-VAE, consisting of $\mathcal{E}$ and $\mathcal{D}$, we efficiently compress optical flow into a low-dimensional latent space. In this section, we provide a brief overview of the Latent Optical Flow Diffusion Model (LOF-DM), which generates optical flows within the latent space. We also discuss the design of the conditioning mechanism and outline the fundamental approach for utilizing these conditions.

\paragraph{Basic architecture.}  
As shown in Fig.~\ref{main}, we visualize the inference phase of LOF-DM. 
The backbone $\epsilon_\theta$ is parametered by $\theta$ and realized by a DiT model~\citep{peebles2023scalable}. 
Overall, we extract the semantic information of $I_1$ and $I_K$ via the CLIP~\citep{radford2021learning} and calculate the linear flow between existing frames as a prior to help the model learning. Notably, this linear flow can be directly computed from the first frame and the last frame, thereby providing a coarse estimation of the ground-truth optical flow.
During the inference phase, the sampling target is $z \sim p_{\theta}(z \mid I_1, I_K)$, where $z$ denote the sampled optical flow in the latent space, and $t$ is uniformly sampled from $\{1, \cdots, T\}$

\paragraph{Existing frames guidance.} 
Interpreting the semantics between the existing frames is important for optical flow generation. In detail, we utilize CLIP to extract semantic information from the given image to achieve image-to-video generation, we apply the same way to extract the useful information. 
Considering the fact that our task requires more fine-grained information, we adapt the ViT~\citep{dosovitskiy2020image} architecture used in CLIP by replacing the global [CLS] token with features derived from individual patch tokens. 
This approach enables us to capture richer and more detailed representations, providing a better foundation for tasks that rely on fine-grained information. 
Similar to CogVideoX~\citep{yang2024cogvideox}, we concatenate the embeddings of both CLIP features and the optical flow in the latent space at the input stage to better align visual and semantic information.

\paragraph{Linear optical flow initialization.}  
Directly generating optical flow from semantic information can be quite challenging, since the complexities and nuances of motion often require a level of detail that pure semantic representations may not capture effectively. Therefore, we consider introducing linear optical flow as a prior to assist this process. While linear optical flow may not be entirely precise, it provides useful information that can help guide the learning model, allowing it to better approximate the underlying motion dynamics and improve overall performance. Formally, we calculate the linear flow as
\begin{equation}
\hat{F}_{k \rightarrow 1}^L = \frac{k-1}{K-1} F_{K \rightarrow 1},
\quad
\hat{F}_{k \rightarrow K}^L = \frac{K-k}{K-1} F_{1 \rightarrow K},
\end{equation}
where $k \in \{2, \ldots, K-1\}$ is the intermediate frame index. These estimated flows are encoded into the latent space using the OF-VAE, expressed as $z_l = \mathcal{E}([\hat{F}_{k \rightarrow 1}^L, \hat{F}_{k \rightarrow K}^L])$. During both the training and inference phases, $z_l$ is concatenated along the channel dimension to support optical flow prediction. Although real-world motions are inherently more complex, we find that initializing with linear motion improves the denoising learning process and accelerates training convergence.

\subsection{Post-Hoc Refinement}
\label{method4.4}

After estimating the optical flow $\hat{F}_{: \rightarrow 1}$ and $\hat{F}_{: \rightarrow K}$ using the OF-VAE and LOF-DM given $I_1$ and $I_K$, we can reconstruct the intermediate frames $\hat{I}_k$ as follows:
\begin{equation}
    \hat{I}_k = \mathcal{P}(\mathcal{W}(I_1, \hat{F}_{k \rightarrow 1}), \mathcal{W}(I_K, \hat{F}_{k \rightarrow K})),
\end{equation}
where $\mathcal{P}$ and $\mathcal{W}$ denotes the reconstruction method and backward warping, respectively.
Previous methods like RIFE use convolutional neural networks to refine the warped results, limiting their ability to generate diverse and detailed content. 
Instead, we propose our MotionControlNet, utilizing the strong video prior learned in current Image-to-Video diffusion models to refine the intermediate frames, leading to better realistic generation. 
Formally, the process is represented as follows:
\begin{equation} 
v \sim p_{\phi}(V \mid I_1, I_K, P, \mathcal{W}(I_1, \hat{F}_{:\rightarrow 1}), \mathcal{W}(I_K, \hat{F}_{:\rightarrow K})), 
\end{equation}
where $P$ and $v$ denote the given text and the final generation frames, respectively. In our experiments, we observe that setting an appropriate prompt can significantly enhance the model's generative capabilities, leading to improved results. 

\paragraph{MotionControlNet.}
Inspired by ControlNet~\citep{zhang2023adding}, which can inject different conditions to existing image diffusion models, we propose the MotionControlNet to generated videos with motion guidance. Formally, we use the CogVideoX-5b-I2V~\citep{yang2024cogvideox} as the basic model and introduce additional trainable \textit{zero convolution layers}. The complete MotionControlNet then computes:

\begin{equation}
\begin{aligned}
    y =& \mathcal{F}_{\phi_1}(I_1, I_K, P\\ &+ \mathcal{Z}_{\phi_2}(I_1, I_K, P, \mathcal{W}(I_1, \hat{F}_{:\rightarrow 1}), \mathcal{W}(I_K, \hat{F}_{:\rightarrow K}))).
\end{aligned}
\end{equation}
where $\mathcal{F}_{\phi_1}$, $\mathcal{Z}_{\phi_2}$ and $ y$ denote the Image-to-Video foundation model parameterized by $\phi_1$, MotionControlNet parameterized by $\phi_2$, and output feature, respectively. For VFI, we use a fixed text prompt $P$, ``Generate a video with smooth motion,'' for all samples. We inject the motion information via the backward warping on key frames, allowing for precise alignment of the generated frames with the motion dynamics of the video. The backward warping process effectively transfers spatial and temporal information from the key frames to the intermediate frames, ensuring smooth transitions and realistic motion generation. By incorporating motion information into the diffusion model, our generated results demonstrate enhanced motion and contextual consistency in comparison to models that generate frames solely using the pair of key frames without integrating motion data. This strategic injection of motion information facilitates a more coherent frame generation process, ultimately leading to superior visual fidelity and continuity in the generated sequences.

\section{Experiment}
\subsection{Experimental setup}
To support large-scale training, we collect video clips from online sources, including Panda-70M~\cite{chen2024panda} and WebVid~\cite{bain2021frozen}. We filter the raw clips using criteria such as aesthetic quality and motion strength, yielding a training set of over $50$ million videos. To accommodate the OF-VAE's $4\times$ temporal compression, we randomly sample training clips at varying spatial resolutions with $4n+1$ frames ($n=1,\ldots,6$), corresponding to $4n\times$ frame-rate conversion, and estimate their optical flow sequences using RAFT~\cite{teed2020raft}. These sequences serve as inputs and reconstruction targets. We build LOF-DM on the CogVideoX-2B architecture~\cite{yang2024cogvideox} and train it on optical flow latents encoded by OF-VAE. For post-hoc refinement, we train MotionControlNet on $500$K samples further selected from the collected data based on resolution and aesthetic scores. All experiments are conducted on NVIDIA H20 GPUs.

\subsection{Results on VFI}
We conduct experiments on the DAVIS-7, UCF101-7~\cite{jain2024video}, and FCVG datasets~\cite{zhu2025generative}. DAVIS-7 and UCF101-7 require interpolating 7 intermediate frames between two input frames, whereas FCVG requires 23 intermediate frames, posing a more challenging interpolation task.

We compare against 4 non-diffusion-based baselines: AMT~\cite{Li_2023_CVPR}, RIFE~\cite{huang2022real}, FILM~\cite{reda2022film}, and GIMM-VFI~\cite{guo2024generalizable}. We also include 7 diffusion-based baselines: LDMVFI~\cite{danier2024ldmvfi}, GI~\cite{wang2025generative}, FCVG~\cite{zhu2025generative}, Wan2.1-Fun-InP-1.3B~\cite{wan2025wan}, ARVFI~\cite{ma2026bi}, MoMo~\cite{lew2025disentangled}, and LDF-VFI~\cite{peng2026towards}. We report numeric comparison results on PSNR, SSIM~\cite{wang2004image}, LPIPS~\cite{zhang2018unreasonable}, FID~\cite{heusel2017gans}, and FVD~\cite{peng2026towards}. 

\begin{table}[t]
    \centering
    \small
    \tabcolsep=0.16cm
    \caption{Quantitative comparison on DAVIS-7. Baseline results with $\dag$ are taken from~\cite{ma2026bi, jain2024video} and a dash (-) indicates that the corresponding metric is not reported for that method in these references. The best results are highlighted in bold.}
    \vskip -0.1in
    \begin{tabular}{lccccc}
        \toprule
        Methods& PSNR $\uparrow$ &SSIM $\uparrow$ & LPIPS$\downarrow$ & FID$\downarrow$ & FVD$\downarrow$ \\
        \midrule
        AMT$\dag$ & \textbf{21.09} & 0.5443 & 0.254 & 34.65 & 234.50 \\
        RIFE$\dag$ & 20.48  &0.5112 & 0.258 & 23.98 & 240.04 \\
        FILM$\dag$  &20.71 &0.5282 & 0.271 & 30.16 & 214.80 \\
        GIMM-VFI$\dag$ &- &- & 0.261  & 27.91 & 227.71 \\
        \midrule
        LDMVFI$\dag$ &19.98 &0.4794 & 0.276 & 22.10 & 245.02 \\
        VIDIM$\dag$ &19.62 &0.4709 &0.258 &28.06 &199.32\\
        GI$\dag$   &- &-   & 0.267 & 27.71 & 211.47 \\
        FCVG$\dag$  &- &-  & 0.266 & 25.96  & 207.17 \\
        Wan$\dag$  &- &-   & 0.323 & 26.97 & 248.14 \\
        ARVFI$\dag$  &- &- & 0.257 & \textbf{21.65} & 188.77 \\
        MoMo & 20.10 &\textbf{0.6533} &0.223 & 23.67&213.10\\
        LDF-VFI &18.23 &0.5800 &0.221 &33.01 &313.03 \\
        \midrule
        LumosFlow & 19.72 &0.5069 &\textbf{0.187} &23.19 &\textbf{157.94}\\
        \bottomrule
    \end{tabular}
    \label{tab:davis-7}
    \vskip -0.15in
\end{table}
\begin{table}[t]
    \centering
    \small
    \caption{Quantitative comparison on UCF101-7. Baseline results are taken from~\cite{jain2024video}. The best results are highlighted in bold.}
    \vskip -0.1in
    \tabcolsep=0.18cm
    \begin{tabular}{lccccc}
        \toprule
        Methods& PSNR $\uparrow$ &SSIM $\uparrow$ & LPIPS$\downarrow$ & FID$\downarrow$ & FVD$\downarrow$ \\
        \midrule
        AMT & \textbf{26.06} & \textbf{0.8139}  & 0.1442 & 31.6 & 344.5 \\
        RIFE & 25.73  &0.8040 & 0.1359 & 18.72 & 323.8 \\
        FILM  &25.90 &0.8118 & 0.1373 & 26.06 & 328.2 \\
        \midrule
        LDMVFI &25.57 &0.8006 &0.1356 &\textbf{18.09} &316.3\\
        VIDIM &24.07 &0.7817 &0.1495 &34.48 &278.0\\
        \midrule
        LumosFlow & 24.52 &0.8081 &\textbf{0.1176} &18.69&\textbf{236.54}\\
        \bottomrule
    \end{tabular}
    \label{tab:ucf101}
    \vskip -0.15in
\end{table}

\begin{table}[t]
    \centering
    \small
    \tabcolsep=0.5cm
    \caption{Quantitative comparison on FCVG. Baseline results are taken from~\cite{ma2026bi}. The best results are highlighted in bold.}
    \vskip -0.1in
    \begin{tabular}{lccc}
        \toprule
        Methods & LPIPS$\downarrow$ & FID$\downarrow$ & FVD$\downarrow$ \\
        \midrule
        AMT  & 0.224 & 44.74 & 375.00 \\
        RIFE & 0.247& 39.01& 366.14 \\
        FILM   & 0.241 &39.82 &279.08 \\
        GIMM-VFI& 0.231 &38.71 & 292.08  \\
        \midrule
        LDMVFI&0.228 & 37.74 &  371.49\\
        GI & 0.334 &  43.08 &  282.22 \\
        FCVG & 0.266 &  31.24 &  225.48 \\
        Wan & 0.223 &  28.52 &  214.45 \\
        ARVFI & 0.206 &\textbf{19.03} & 201.47 \\
        \midrule
        LumosFlow &\textbf{0.153} &22.31&\textbf{176.65}\\
        \bottomrule
    \end{tabular}
    \label{tab:FCVG}
    \vskip -0.2in
\end{table}

\paragraph{Overall results}
As present in Tabs.~\ref{tab:davis-7}, \ref{tab:ucf101}, and \ref{tab:FCVG}, LumosFlow consistently achieves superior performance, securing the best LPIPS and FVD scores across all three evaluation benchmarks. Compared with traditional non-diffusion methods, LumosFlow significantly enhances perceptual quality at both the frame and video levels, reducing FVD by 26.5\% on DAVIS-7 and 26.9\% on UCF101-7 relative to the best traditional baselines on each dataset. Against diffusion-based counterparts, LumosFlow further drives down FVD to 157.94, 236.54, and 176.65 across DAVIS-7, UCF101-7, and FCVG, outperforming state-of-the-art diffusion models like ARVFI and VIDIM. These robust and consistent FVD reductions underscore superior video quality and enhanced temporal coherence. Furthermore, the lowest LPIPS scores achieved by LumosFlow across all datasets (0.187, 0.1176, and 0.153) confirm that these video-level temporal gains are consistently complemented by superior frame-level perceptual fidelity.

\begin{figure}[t]
    \centering
    \includegraphics[width=.9\linewidth]{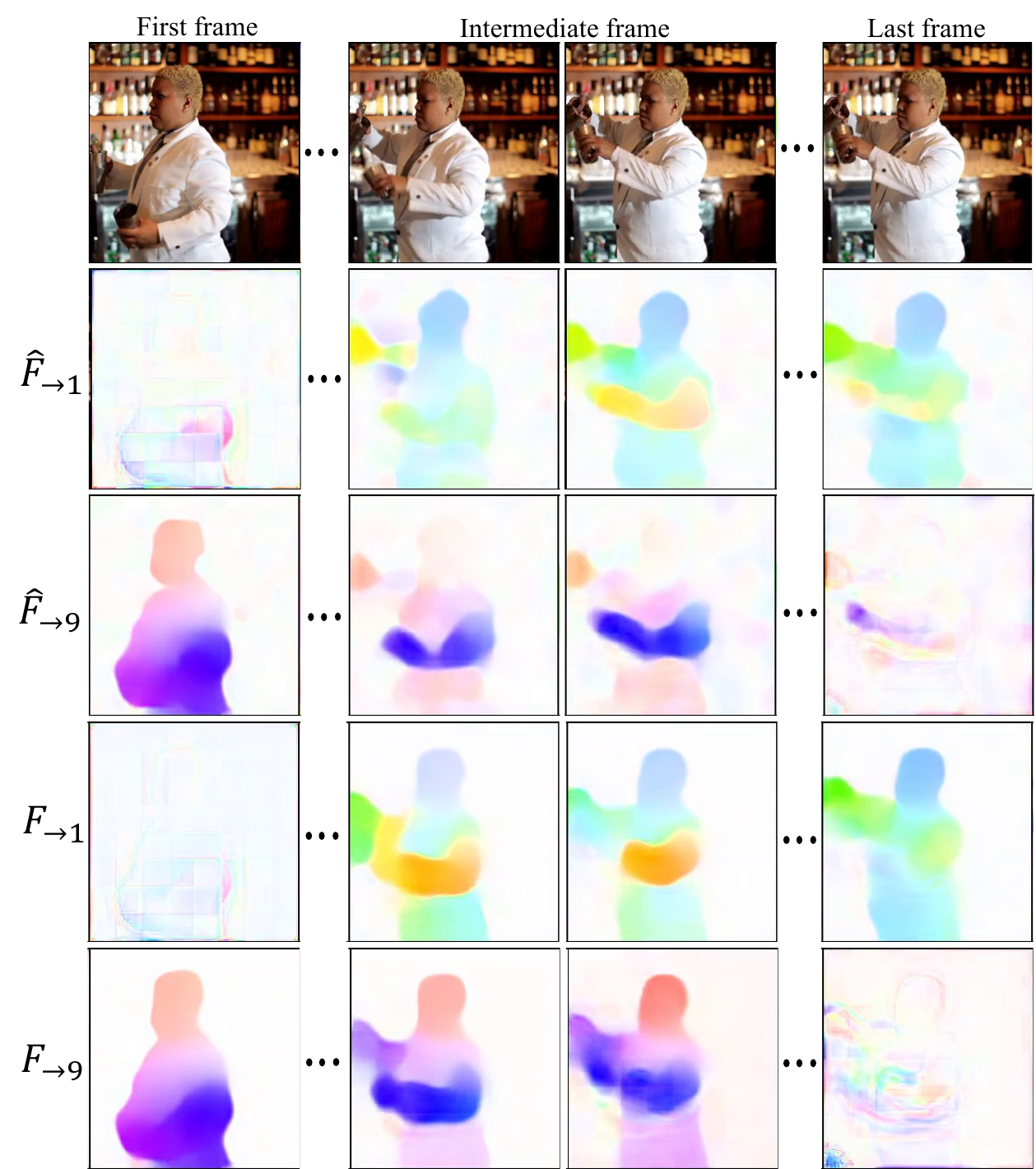}
    \vskip -0.15in
    \caption{Visualization of optical flow generation conditioned on the first and last frames. $\hat{F}_{\rightarrow 1}$ and $\hat{F}_{\rightarrow 9}$ denote the generated optical flows to the first and last frames, respectively, while $F_{\rightarrow 1}$ and $F_{\rightarrow 9}$ denote the corresponding flows reconstructed by OF-VAE.
    }
    \label{flow_vis}
    \vskip -0.15in
\end{figure}

\begin{figure*}
    \centering
    \includegraphics[width=0.95\linewidth]{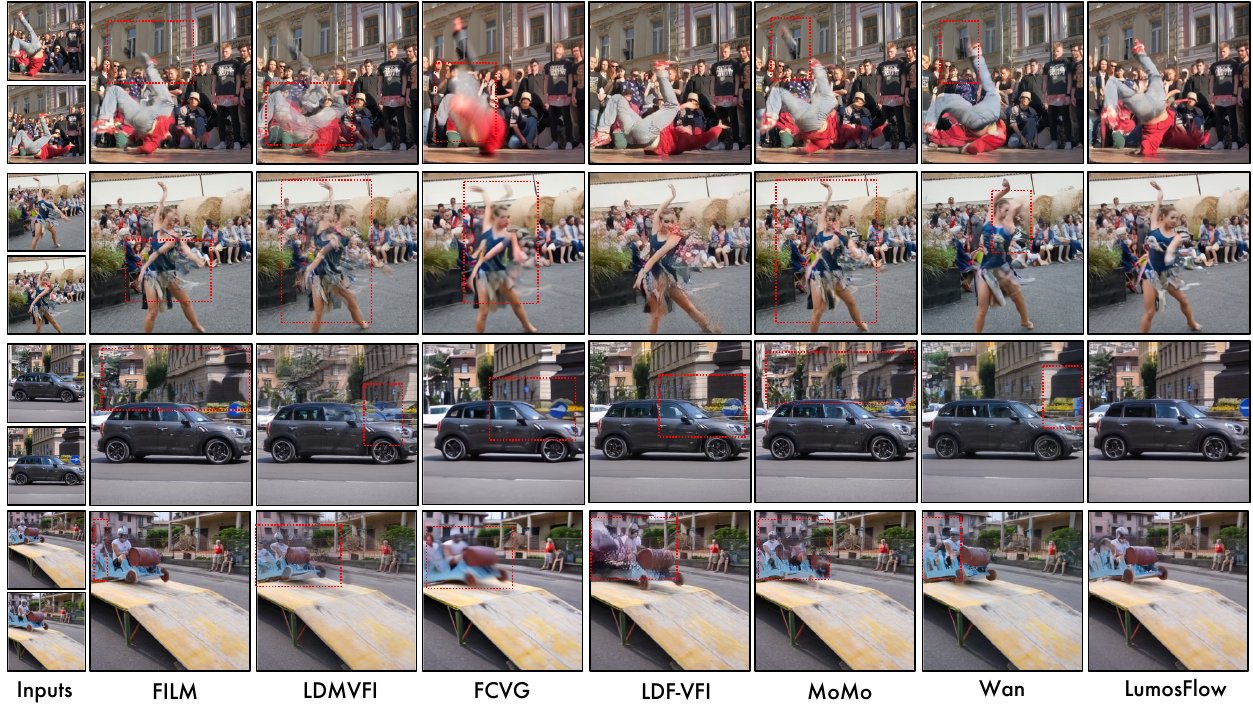}
    \vskip -0.15in
    \caption{Visual comparison of mid-frames across different methods. We highlight artifacts with red boxes.}
    \label{fig:vis_vfi}
    \vskip -0.2in
\end{figure*}

\begin{table}[t]
    \centering
    \small
    \tabcolsep=0.15cm
    \caption{The ablation study on motion guidance on DAVIS-7.}
    \vskip -0.1in
    \begin{tabular}{lccccc}
        \toprule
        & PSNR $\uparrow$ &SSIM $\uparrow$ & LPIPS$\downarrow$ & FID$\downarrow$ & FVD$\downarrow$ \\
        \midrule
        w/o motion &19.41 &0.4945 &0.1949 &22.48 &162.19\\
        LumosFlow & 19.72 &0.5069 &\textbf{0.187} &23.19 &\textbf{157.94}\\
        \bottomrule
    \end{tabular}
    \label{tab:motion_guidance}
    \vskip -0.15in
\end{table}
\begin{table}[t]
    \centering
    \small
    \tabcolsep=0.13cm
    \caption{EPE between generated and RAFT-estimated flows. The shaded OF-VAE row reports reconstruction error after compression. Avg and Mid denote the average over all intermediate frames and the midpoint result, respectively; First and Last indicate flows to the corresponding input frames.}
    \vskip -0.1in
    \begin{tabular}{l|c|c|c|c}
        \toprule
         & Avg (First) & Avg (Last) & Mid (First) & Mid(Last)\\
        \midrule
        \rowcolor[gray]{0.8}
        OF-VAE  & 0.520 &0.592 &0.674 & 0.683 \\
        \midrule
        Linear flow &1.342 & 1.297 &1.477 &1.493  \\
        MoMo & -  & - & 1.486 &1.497\\
        \midrule
        LOF-DM & \textbf{1.306} &\textbf{1.243} &\textbf{1.362} &\textbf{1.375}\\
        \bottomrule
    \end{tabular}
    \vskip -0.1in
    \label{flow_gen_ab}
\end{table}
\begin{table}[t]
    \centering
    \small
    \tabcolsep=0.08cm
    \caption{Experimental results on different evaluation metrics. }
    \vskip -0.1in
    \begin{tabular}{l|c|c|c|c|c|c}
        \toprule
         & FVD $\downarrow$ & FID $\downarrow$  & S-C $\uparrow$ & M-S $\uparrow$ & T-F $\uparrow$ & D-D$\uparrow$ \\
         \midrule
         FreeLong& 1829.30 & 482.47 &\textbf{0.9843} &0.9752 & 0.9565 & 0.2175\\
         FreeNoise& 2175.56 &481.69  & 0.9778 & 0.9722 & 0.9672 & 0.3142\\
         Video-Infinity &1788.95 &483.33  &0.9342 &0.9566 &0.9542 &0.5470\\
         \midrule
         Ours &\textbf{912.83} &\textbf{479.56} &0.9541 &\textbf{0.9898} &\textbf{0.9762} &\textbf{0.5700}\\
         \bottomrule
    \end{tabular}
    \vskip -0.15in
    \label{longvideo_res}
\end{table}

\begin{table}[t]
    \centering
    \small
    \caption{Human study of video generation.}   
    \tabcolsep=0.26cm
    \vskip -0.1in
    \begin{tabular}{l|c|c|c|c|c}
    \toprule
         &  T-A & F-C & D-D & V-Q & P-R\\
         \midrule
        FreeLong & 2.18 &2.73 &1.92 &2.07 & 0.85\%\\
        FreeNoise &2.37 &2.03 &2.52 &1.83 & 0.00\%\\
        Video-Infinity & 2.44 &2.11 &2.46 &1.88 & 0.85\%\\
        \midrule
        Ours & \textbf{4.14} & \textbf{3.69} &\textbf{3.44} & \textbf{3.91} & \textbf{98.2\%}\\
        \bottomrule
    \end{tabular}
    \label{tab_human_study}
    \vskip -0.2in
\end{table}
\paragraph{Visualization of VFI}
We select a video and extract the first and last frames as a pair of key frames to generate the possible motion between them, as shown in Fig.~\ref{flow_vis}. For the given sample, it can be clearly seen that LOF-DM models the movement of the arm. Moreover, the direction of the arm's movement has changed, indicating the difference between our generated optical flow and linear optical flow.

In addition, LumosFlow synthesizes intermediate frames with fewer artifacts and better preserves human body structures and object boundaries, as shown in Fig.~\ref{fig:vis_vfi}. Other methods exhibit noticeable artifacts, including distorted limbs, ghosting, and blurred details, as highlighted by the red boxes. Although LDF-VFI produces relatively plausible individual frames in some cases, its intermediate predictions often closely resemble the input frames, failing to capture the expected motion progression. Thus, plausible frame appearance alone does not ensure accurate interpolation. LumosFlow better captures intermediate motion while maintaining visual fidelity, producing more coherent transitions between the input frames. 

\begin{figure*}[t]
    \centering
    \includegraphics[width=0.9\linewidth]{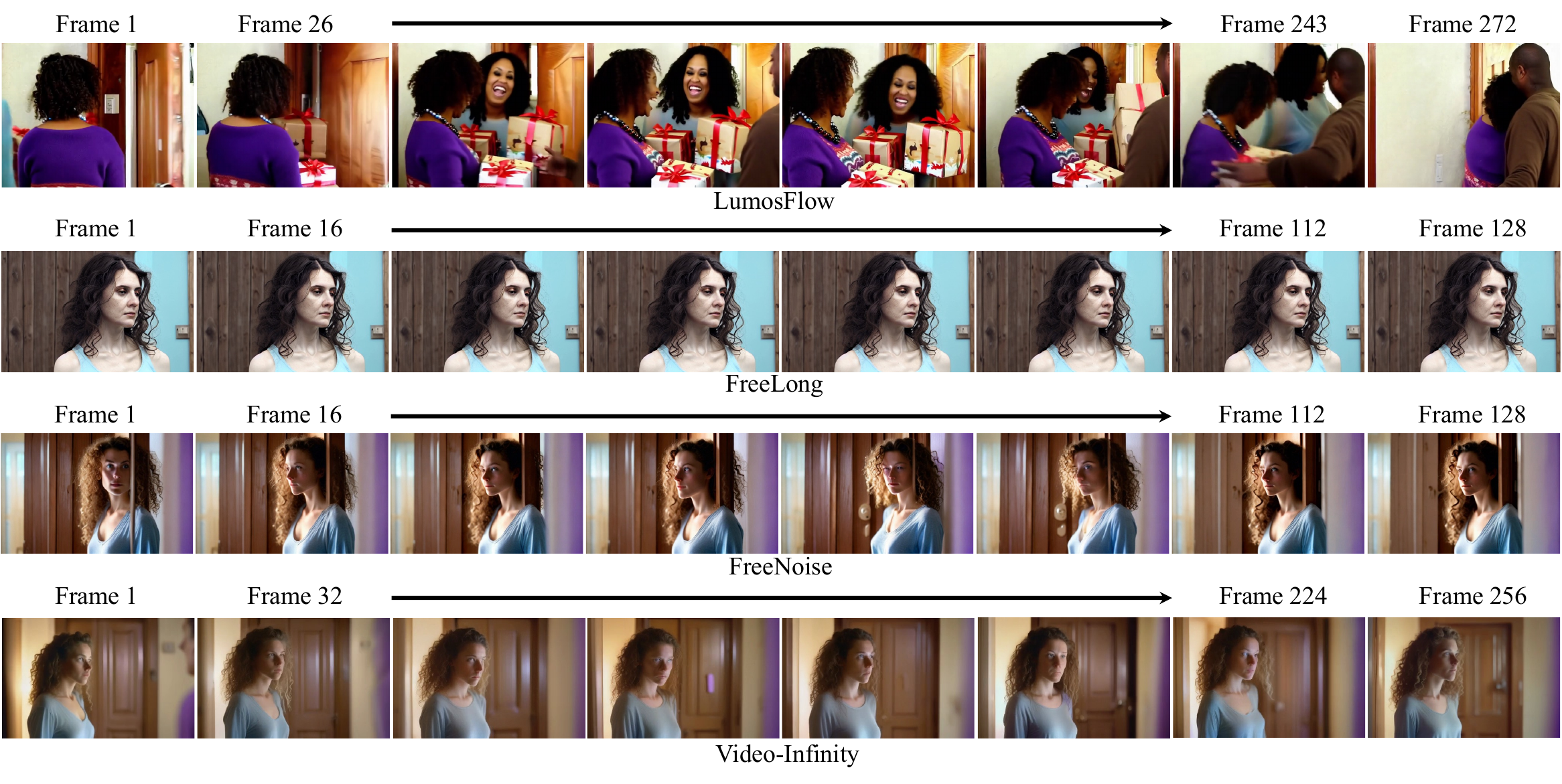}
    \vskip -0.15in
    \caption{The generated long videos via LumosFlow, FreeLong, FreeNoise, Video-Infinity by given the prompt ``A woman with long, wavy hair tied back, wearing a sheer, light blue top with thin strap. In the warm interior, a curly-haired woman..."}
    \label{long_res}
    \vskip -0.2in
\end{figure*}
\paragraph{Ablation study on motion guidance}
To evaluate the importance of motion guidance, we modify CogVideoX-I2V to condition on the first and last frames and directly generate intermediate frames. We train this baseline using the same data and training procedure as MotionControlNet. As shown in Tab.~\ref{tab:motion_guidance}, introducing motion guidance improves both PSNR and SSIM, while reducing LPIPS from 0.1949 to 0.187 and FVD from 162.19 to 157.94. These results demonstrate the benefit of explicit motion guidance in improving frame fidelity and video-level perceptual quality.

\paragraph{Quantitative results on motion generation}
We evaluate motion generation conditioned on the first and last frames using the endpoint error (EPE) between generated and RAFT-estimated flows. For reference, we report the reconstruction EPE of OF-VAE to quantify the error introduced by compression. We compare LOF-DM with linear flow interpolation across all intermediate frames and with MoMo only at the temporal midpoint, consistent with its design for midpoint interpolation. As shown in Tab.~\ref{flow_gen_ab}, LOF-DM achieves lower average EPE than linear interpolation and outperforms both baselines at the midpoint, for flows to both endpoints. For nine flow frames at $256\times256$, a single LOF-DM denoising forward pass takes about $0.1$ seconds on one H20 GPU (batch size 1, half precision, without CFG). Additional evaluations of different compression ratios are provided in Appendix Sec.~H.

\subsection{Results on video generation}
Beyond VFI, we also apply LumosFlow to video generation by synthesizing intermediate frames between successive keyframes. We first fine-tune CogVideoX-5B-T2V on low-frame-rate videos, enabling it to generate keyframes separated by large temporal intervals. At inference time, given a text prompt, the model generates 18 sparse keyframes at a resolution of $480\times640$. LumosFlow then generates 15 intermediate frames between each consecutive pair, producing a 273-frame video. More details about the key frame generation are in Appendix Sec. B and C.

\paragraph{Quantitative comparison} We compare LumosFlow with other video generation methods FreeLong~\cite{lu2025freelong}, FreeNoise~\cite{qiu2023freenoise}, and Video-Infinity~\cite{tan2024videoinf} and report FVD, FID, Subject Consistency (S-C), Motion Smoothness (M-S), Temporal Flickering (T-F), and Dynamic Degree (D-D) following VBench~\cite{huang2023vbench}. For a fair comparison, we construct a small test set containing 100 high-quality text-video pairs and apply these methods to generate videos corresponding to these texts. As shown in Tab.~\ref{longvideo_res}, LumosFlow presents outstanding performance in FVD and FID, suggesting that it effectively generates high-quality and diverse video content. Moreover, LumosFlow achieves the best performance in Motion Smoothness and Dynamic Degree, indicating that our method not only ensures smooth transitions but also captures a high level of diversity in the generated video sequences. Although FreeLong and FreeNoise perform well in Subject Consistency, their lower Dynamic Degree scores suggest more limited motion, highlighting the advantage of LumosFlow in generating visually coherent and dynamic videos.

\paragraph{Human study}
For human study, we collect $14$ videos and consider Text Alignment (T-A), Frame Consistency (F-C), Dynamic Degree (D-D), and Video Quality (V-Q) metrics, ranging from $1$ (very low) to $5$ (very large). In addition, all users are required to choose the best video generated by different methods, referring to Preference Rate (P-R). As shown in Tab.~\ref{tab_human_study}, LumosFlow achieves the best performance among all the metrics. Specifically, LumosFlow achieves the highest score of Dynamic Degree and Video Quality at the same time, indicating the generated video has large frame-to-frame differences, but the overall quality is high.

\paragraph{Visualization results}
As shown in Fig.~\ref{long_res},  we generate videos based on a specific prompt using LumosFlow, FreeLong, FreeNoise, and Video-Infinity. Compared to the other methods, LumosFlow demonstrates significant movement between the various frames, indicating dynamic activity throughout the sequence. Despite this motion, the core elements of the video are consistently maintained, ensuring visual coherence. In contrast, the videos generated by the other methods exhibit temporal repetition, existing minimal motion across different frames. Additional generated videos for each method can be found in Appendix Sec. G.
\section{Conclusion}
We present LumosFlow, which generates optical flow in a compact latent space to guide frame refinement. Motivated by our compressibility analysis, it enables efficient training and motion generation, achieving temporally coherent interpolation under large, complex motion and supporting keyframe-based video generation. Future work will explore extensions to longer sequences and video editing.

{
    \small
    \bibliographystyle{ieeenat_fullname}
    \bibliography{main}

\begin{thebibliography}{46}
\providecommand{\natexlab}[1]{#1}
\providecommand{\url}[1]{\texttt{#1}}
\expandafter\ifx\csname urlstyle\endcsname\relax
  \providecommand{\doi}[1]{doi: #1}\else
  \providecommand{\doi}{doi: \begingroup \urlstyle{rm}\Url}\fi

\bibitem[aigc apps(2026)]{aigc_apps_VideoX_Fun_2026}
aigc apps.
\newblock Videox-fun: A video generation pipeline for diffusion transformer, 2026.

\bibitem[Bain et~al.(2021)Bain, Nagrani, Varol, and Zisserman]{bain2021frozen}
Max Bain, Arsha Nagrani, G{\"u}l Varol, and Andrew Zisserman.
\newblock Frozen in time: A joint video and image encoder for end-to-end retrieval.
\newblock In \emph{2021 IEEE/CVF International Conference on Computer Vision (ICCV)}, pages 1708--1718. IEEE, 2021.

\bibitem[Bao et~al.(2019)Bao, Lai, Ma, Zhang, Gao, and Yang]{bao2019depth}
Wenbo Bao, Wei-Sheng Lai, Chao Ma, Xiaoyun Zhang, Zhiyong Gao, and Ming-Hsuan Yang.
\newblock Depth-aware video frame interpolation.
\newblock In \emph{Proceedings of the IEEE/CVF conference on computer vision and pattern recognition}, pages 3703--3712, 2019.

\bibitem[Blattmann et~al.(2023)Blattmann, Dockhorn, Kulal, Mendelevitch, Kilian, Lorenz, Levi, English, Voleti, Letts, et~al.]{blattmann2023stable}
Andreas Blattmann, Tim Dockhorn, Sumith Kulal, Daniel Mendelevitch, Maciej Kilian, Dominik Lorenz, Yam Levi, Zion English, Vikram Voleti, Adam Letts, et~al.
\newblock Stable video diffusion: Scaling latent video diffusion models to large datasets.
\newblock \emph{arXiv preprint arXiv:2311.15127}, 2023.

\bibitem[Chen et~al.(2025)Chen, Ma, Liu, Li, Chen, Liu, He, Li, He, and Wu]{chen2025humo}
Liyang Chen, Tianxiang Ma, Jiawei Liu, Bingchuan Li, Zhuowei Chen, Lijie Liu, Xu He, Gen Li, Qian He, and Zhiyong Wu.
\newblock Humo: Human-centric video generation via collaborative multi-modal conditioning.
\newblock \emph{arXiv preprint arXiv:2509.08519}, 2025.

\bibitem[Chen et~al.(2024)Chen, Siarohin, Menapace, Deyneka, Chao, Jeon, Fang, Lee, Ren, Yang, et~al.]{chen2024panda}
Tsai-Shien Chen, Aliaksandr Siarohin, Willi Menapace, Ekaterina Deyneka, Hsiang-wei Chao, Byung~Eun Jeon, Yuwei Fang, Hsin-Ying Lee, Jian Ren, Ming-Hsuan Yang, et~al.
\newblock Panda-70m: Captioning 70m videos with multiple cross-modality teachers.
\newblock In \emph{2024 IEEE/CVF Conference on Computer Vision and Pattern Recognition (CVPR)}, pages 13320--13331. IEEE, 2024.

\bibitem[Chen et~al.(2023)Chen, Ji, Wu, Wu, Xie, Li, Xia, Xiao, and Lin]{chen2023control}
Weifeng Chen, Yatai Ji, Jie Wu, Hefeng Wu, Pan Xie, Jiashi Li, Xin Xia, Xuefeng Xiao, and Liang Lin.
\newblock Control-a-video: Controllable text-to-video diffusion models with motion prior and reward feedback learning.
\newblock \emph{arXiv preprint arXiv:2305.13840}, 2023.

\bibitem[Danier et~al.(2024)Danier, Zhang, and Bull]{danier2024ldmvfi}
Duolikun Danier, Fan Zhang, and David Bull.
\newblock Ldmvfi: Video frame interpolation with latent diffusion models.
\newblock In \emph{Proceedings of the AAAI Conference on Artificial Intelligence}, pages 1472--1480, 2024.

\bibitem[Dosovitskiy(2020)]{dosovitskiy2020image}
Alexey Dosovitskiy.
\newblock An image is worth 16x16 words: Transformers for image recognition at scale.
\newblock \emph{arXiv preprint arXiv:2010.11929}, 2020.

\bibitem[Guo et~al.(2024)Guo, Li, and Loy]{guo2024generalizable}
Zujin Guo, Wei Li, and Chen~Change Loy.
\newblock Generalizable implicit motion modeling for video frame interpolation.
\newblock \emph{Advances in neural information processing systems}, 37:\penalty0 63747--63770, 2024.

\bibitem[HaCohen et~al.(2024)HaCohen, Chiprut, Brazowski, Shalem, Moshe, Richardson, Levin, Shiran, Zabari, Gordon, et~al.]{hacohen2024ltx}
Yoav HaCohen, Nisan Chiprut, Benny Brazowski, Daniel Shalem, Dudu Moshe, Eitan Richardson, Eran Levin, Guy Shiran, Nir Zabari, Ori Gordon, et~al.
\newblock Ltx-video: Realtime video latent diffusion.
\newblock \emph{arXiv preprint arXiv:2501.00103}, 2024.

\bibitem[Hai et~al.(2025)Hai, Wang, Su, Jiang, and Hu]{hai2025hierarchical}
Yang Hai, Guo Wang, Tan Su, Wenjie Jiang, and Yinlin Hu.
\newblock Hierarchical flow diffusion for efficient frame interpolation.
\newblock In \emph{2025 IEEE/CVF Conference on Computer Vision and Pattern Recognition (CVPR)}, pages 22943--22952. IEEE, 2025.

\bibitem[Heusel et~al.(2017)Heusel, Ramsauer, Unterthiner, Nessler, and Hochreiter]{heusel2017gans}
Martin Heusel, Hubert Ramsauer, Thomas Unterthiner, Bernhard Nessler, and Sepp Hochreiter.
\newblock Gans trained by a two time-scale update rule converge to a local nash equilibrium.
\newblock \emph{Advances in neural information processing systems}, 30, 2017.

\bibitem[Ho et~al.(2020)Ho, Jain, and Abbeel]{ho2020denoising}
Jonathan Ho, Ajay Jain, and Pieter Abbeel.
\newblock Denoising diffusion probabilistic models.
\newblock \emph{Advances in neural information processing systems}, 33:\penalty0 6840--6851, 2020.

\bibitem[Huang et~al.(2022)Huang, Zhang, Heng, Shi, and Zhou]{huang2022real}
Zhewei Huang, Tianyuan Zhang, Wen Heng, Boxin Shi, and Shuchang Zhou.
\newblock Real-time intermediate flow estimation for video frame interpolation.
\newblock In \emph{European Conference on Computer Vision}, pages 624--642. Springer, 2022.

\bibitem[Huang et~al.(2024)Huang, He, Yu, Zhang, Si, Jiang, Zhang, Wu, Jin, Chanpaisit, Wang, Chen, Wang, Lin, Qiao, and Liu]{huang2023vbench}
Ziqi Huang, Yinan He, Jiashuo Yu, Fan Zhang, Chenyang Si, Yuming Jiang, Yuanhan Zhang, Tianxing Wu, Qingyang Jin, Nattapol Chanpaisit, Yaohui Wang, Xinyuan Chen, Limin Wang, Dahua Lin, Yu Qiao, and Ziwei Liu.
\newblock {VBench}: Comprehensive benchmark suite for video generative models.
\newblock In \emph{Proceedings of the IEEE/CVF Conference on Computer Vision and Pattern Recognition}, 2024.

\bibitem[Jain et~al.(2024)Jain, Watson, Tabellion, Poole, Kontkanen, et~al.]{jain2024video}
Siddhant Jain, Daniel Watson, Eric Tabellion, Ben Poole, Janne Kontkanen, et~al.
\newblock Video interpolation with diffusion models.
\newblock In \emph{Proceedings of the IEEE/CVF Conference on Computer Vision and Pattern Recognition}, pages 7341--7351, 2024.

\bibitem[Kong et~al.(2024)Kong, Tian, Zhang, Min, Dai, Zhou, Xiong, Li, Wu, Zhang, et~al.]{kong2024hunyuanvideo}
Weijie Kong, Qi Tian, Zijian Zhang, Rox Min, Zuozhuo Dai, Jin Zhou, Jiangfeng Xiong, Xin Li, Bo Wu, Jianwei Zhang, et~al.
\newblock Hunyuanvideo: A systematic framework for large video generative models.
\newblock \emph{arXiv preprint arXiv:2412.03603}, 2024.

\bibitem[Kullback and Leibler(1951)]{kullback1951information}
Solomon Kullback and Richard~A Leibler.
\newblock On information and sufficiency.
\newblock \emph{The annals of mathematical statistics}, 22\penalty0 (1):\penalty0 79--86, 1951.

\bibitem[Lew et~al.(2025)Lew, Choi, Shin, Jung, and Yoon]{lew2025disentangled}
Jaihyun Lew, Jooyoung Choi, Chaehun Shin, Dahuin Jung, and Sungroh Yoon.
\newblock Disentangled motion modeling for video frame interpolation.
\newblock In \emph{Proceedings of the AAAI Conference on Artificial Intelligence}, pages 4607--4615, 2025.

\bibitem[Li et~al.(2023)Li, Zhu, Han, Hou, Guo, and Cheng]{Li_2023_CVPR}
Zhen Li, Zuo-Liang Zhu, Ling-Hao Han, Qibin Hou, Chun-Le Guo, and Ming-Ming Cheng.
\newblock Amt: All-pairs multi-field transforms for efficient frame interpolation.
\newblock In \emph{Proceedings of the IEEE/CVF Conference on Computer Vision and Pattern Recognition (CVPR)}, pages 9801--9810, 2023.

\bibitem[Liu et~al.(2017)Liu, Yeh, Tang, Liu, and Agarwala]{liu2017video}
Ziwei Liu, Raymond~A Yeh, Xiaoou Tang, Yiming Liu, and Aseem Agarwala.
\newblock Video frame synthesis using deep voxel flow.
\newblock In \emph{Proceedings of the IEEE international conference on computer vision}, pages 4463--4471, 2017.

\bibitem[Lu et~al.(2024)Lu, Liang, Zhu, and Yang]{lu2025freelong}
Yu Lu, Yuanzhi Liang, Linchao Zhu, and Yi Yang.
\newblock Freelong: Training-free long video generation with spectralblend temporal attention.
\newblock In \emph{Advances in Neural Information Processing Systems}, pages 131434--131455. Curran Associates, Inc., 2024.

\bibitem[Ma et~al.(2026{\natexlab{a}})Ma, Zhao, Yao, Li, Liu, Dou, Gu, Xue, et~al.]{ma2026realtime}
Yongrui Ma, Shijie Zhao, Mingde Yao, Junlin Li, Xiaohong Liu, Qi Dou, Jinwei Gu, Tianfan Xue, et~al.
\newblock Realtime video frame interpolation using one-step diffusion sampling.
\newblock In \emph{The Fourteenth International Conference on Learning Representations}, 2026{\natexlab{a}}.

\bibitem[Ma et~al.(2026{\natexlab{b}})Ma, Zhao, Yao, Li, Zhang, Liu, Dou, Gu, and Xue]{ma2026bi}
Yongrui Ma, Shijie Zhao, Mingde Yao, Junlin Li, Li Zhang, Xiaohong Liu, Qi Dou, Jinwei Gu, and Tianfan Xue.
\newblock Bi-directional autoregressive diffusion for large complex motion interpolation.
\newblock In \emph{Proceedings of the IEEE/CVF Conference on Computer Vision and Pattern Recognition}, pages 35779--35788, 2026{\natexlab{b}}.

\bibitem[Niklaus et~al.(2017)Niklaus, Mai, and Liu]{niklaus2017video}
Simon Niklaus, Long Mai, and Feng Liu.
\newblock Video frame interpolation via adaptive separable convolution.
\newblock In \emph{Proceedings of the IEEE international conference on computer vision}, pages 261--270, 2017.

\bibitem[Peebles and Xie(2023)]{peebles2023scalable}
William Peebles and Saining Xie.
\newblock Scalable diffusion models with transformers.
\newblock In \emph{Proceedings of the IEEE/CVF International Conference on Computer Vision}, pages 4195--4205, 2023.

\bibitem[Peng et~al.(2026)Peng, Li, Huang, Zheng, Wang, Chen, Dai, Li, Zou, and Xiong]{peng2026towards}
Xinyu Peng, Han Li, Yuyang Huang, Ziyang Zheng, Yaoming Wang, Xin Chen, Wenrui Dai, Chenglin Li, Junni Zou, and Hongkai Xiong.
\newblock Towards holistic modeling for video frame interpolation with auto-regressive diffusion transformers.
\newblock \emph{arXiv preprint arXiv:2601.14959}, 2026.

\bibitem[Podell et~al.(2023)Podell, English, Lacey, Blattmann, Dockhorn, M{\"u}ller, Penna, and Rombach]{podell2023sdxl}
Dustin Podell, Zion English, Kyle Lacey, Andreas Blattmann, Tim Dockhorn, Jonas M{\"u}ller, Joe Penna, and Robin Rombach.
\newblock Sdxl: Improving latent diffusion models for high-resolution image synthesis.
\newblock \emph{arXiv preprint arXiv:2307.01952}, 2023.

\bibitem[Qiu et~al.(2023)Qiu, Xia, Zhang, He, Wang, Shan, and Liu]{qiu2023freenoise}
Haonan Qiu, Menghan Xia, Yong Zhang, Yingqing He, Xintao Wang, Ying Shan, and Ziwei Liu.
\newblock Freenoise: Tuning-free longer video diffusion via noise rescheduling.
\newblock \emph{arXiv preprint arXiv:2310.15169}, 2023.

\bibitem[Radford et~al.(2021)Radford, Kim, Hallacy, Ramesh, Goh, Agarwal, Sastry, Askell, Mishkin, Clark, et~al.]{radford2021learning}
Alec Radford, Jong~Wook Kim, Chris Hallacy, Aditya Ramesh, Gabriel Goh, Sandhini Agarwal, Girish Sastry, Amanda Askell, Pamela Mishkin, Jack Clark, et~al.
\newblock Learning transferable visual models from natural language supervision.
\newblock In \emph{International conference on machine learning}, pages 8748--8763. PMLR, 2021.

\bibitem[Reda et~al.(2022)Reda, Kontkanen, Tabellion, Sun, Pantofaru, and Curless]{reda2022film}
Fitsum Reda, Janne Kontkanen, Eric Tabellion, Deqing Sun, Caroline Pantofaru, and Brian Curless.
\newblock Film: Frame interpolation for large motion.
\newblock In \emph{European Conference on Computer Vision}, pages 250--266. Springer, 2022.

\bibitem[Rombach et~al.(2022)Rombach, Blattmann, Lorenz, Esser, and Ommer]{rombach2022high}
Robin Rombach, Andreas Blattmann, Dominik Lorenz, Patrick Esser, and Bj{\"o}rn Ommer.
\newblock High-resolution image synthesis with latent diffusion models.
\newblock In \emph{Proceedings of the IEEE/CVF conference on computer vision and pattern recognition}, pages 10684--10695, 2022.

\bibitem[Tan et~al.(2024)Tan, Yang, Liu, and Wang]{tan2024videoinf}
Zhenxiong Tan, Xingyi Yang, Songhua Liu, and Xinchao Wang.
\newblock Video-infinity: Distributed long video generation.
\newblock \emph{arXiv preprint arXiv:2406.16260}, 2024.

\bibitem[Teed and Deng(2020)]{teed2020raft}
Zachary Teed and Jia Deng.
\newblock Raft: Recurrent all-pairs field transforms for optical flow.
\newblock In \emph{Computer Vision--ECCV 2020: 16th European Conference, Glasgow, UK, August 23--28, 2020, Proceedings, Part II 16}, pages 402--419. Springer, 2020.

\bibitem[Wan et~al.(2025)Wan, Wang, Ai, Wen, Mao, Xie, Chen, Yu, Zhao, Yang, et~al.]{wan2025wan}
Team Wan, Ang Wang, Baole Ai, Bin Wen, Chaojie Mao, Chen-Wei Xie, Di Chen, Feiwu Yu, Haiming Zhao, Jianxiao Yang, et~al.
\newblock Wan: Open and advanced large-scale video generative models.
\newblock \emph{arXiv preprint arXiv:2503.20314}, 2025.

\bibitem[Wang et~al.(2023)Wang, Yuan, Zhang, Chen, Wang, Zhang, Shen, Zhao, and Zhou]{wang2023videocomposer}
Xiang Wang, Hangjie Yuan, Shiwei Zhang, Dayou Chen, Jiuniu Wang, Yingya Zhang, Yujun Shen, Deli Zhao, and Jingren Zhou.
\newblock Videocomposer: Compositional video synthesis with motion controllability.
\newblock \emph{Advances in Neural Information Processing Systems}, 36:\penalty0 7594--7611, 2023.

\bibitem[Wang et~al.(2025)Wang, Zhou, Curless, Kemelmacher-Shlizerman, Holynski, and Seitz]{wang2025generative}
Xiaojuan Wang, Boyang Zhou, Brian Curless, Ira Kemelmacher-Shlizerman, Aleksander Holynski, and Steve Seitz.
\newblock Generative inbetweening: Adapting image-to-video models for keyframe interpolation.
\newblock In \emph{International Conference on Learning Representations}, pages 30594--30609, 2025.

\bibitem[Wang et~al.(2004)Wang, Bovik, Sheikh, and Simoncelli]{wang2004image}
Zhou Wang, Alan~C Bovik, Hamid~R Sheikh, and Eero~P Simoncelli.
\newblock Image quality assessment: from error visibility to structural similarity.
\newblock \emph{IEEE transactions on image processing}, 13\penalty0 (4):\penalty0 600--612, 2004.

\bibitem[Xu et~al.(2019)Xu, Siyao, Sun, Yin, and Yang]{xu2019quadratic}
Xiangyu Xu, Li Siyao, Wenxiu Sun, Qian Yin, and Ming-Hsuan Yang.
\newblock Quadratic video interpolation.
\newblock \emph{Advances in Neural Information Processing Systems}, 32, 2019.

\bibitem[Yang et~al.(2024)Yang, Teng, Zheng, Ding, Huang, Xu, Yang, Hong, Zhang, Feng, et~al.]{yang2024cogvideox}
Zhuoyi Yang, Jiayan Teng, Wendi Zheng, Ming Ding, Shiyu Huang, Jiazheng Xu, Yuanming Yang, Wenyi Hong, Xiaohan Zhang, Guanyu Feng, et~al.
\newblock Cogvideox: Text-to-video diffusion models with an expert transformer.
\newblock \emph{arXiv preprint arXiv:2408.06072}, 2024.

\bibitem[Yin et~al.(2023)Yin, Wu, Liang, Shi, Li, Ming, and Duan]{yin2023dragnuwa}
Shengming Yin, Chenfei Wu, Jian Liang, Jie Shi, Houqiang Li, Gong Ming, and Nan Duan.
\newblock Dragnuwa: Fine-grained control in video generation by integrating text, image, and trajectory.
\newblock \emph{arXiv preprint arXiv:2308.08089}, 2023.

\bibitem[Zhang et~al.(2023)Zhang, Rao, and Agrawala]{zhang2023adding}
Lvmin Zhang, Anyi Rao, and Maneesh Agrawala.
\newblock Adding conditional control to text-to-image diffusion models.
\newblock In \emph{Proceedings of the IEEE/CVF International Conference on Computer Vision}, pages 3836--3847, 2023.

\bibitem[Zhang et~al.(2018)Zhang, Isola, Efros, Shechtman, and Wang]{zhang2018unreasonable}
Richard Zhang, Phillip Isola, Alexei~A Efros, Eli Shechtman, and Oliver Wang.
\newblock The unreasonable effectiveness of deep features as a perceptual metric.
\newblock In \emph{2018 IEEE/CVF conference on computer vision and pattern recognition}, pages 586--595. IEEE, 2018.

\bibitem[Zhang et~al.(2025)Zhang, Chen, Zhao, Lu, Fu, Xu, and Wu]{zhang2025eden}
Zihao Zhang, Haoran Chen, Haoyu Zhao, Guansong Lu, Yanwei Fu, Hang Xu, and Zuxuan Wu.
\newblock Eden: Enhanced diffusion for high-quality large-motion video frame interpolation.
\newblock In \emph{2025 IEEE/CVF Conference on Computer Vision and Pattern Recognition (CVPR)}, pages 2105--2115. IEEE, 2025.

\bibitem[Zhu et~al.(2025)Zhu, Ren, Wang, Wu, and Zuo]{zhu2025generative}
Tianyi Zhu, Dongwei Ren, Qilong Wang, Xiaohe Wu, and Wangmeng Zuo.
\newblock Generative inbetweening through frame-wise conditions-driven video generation.
\newblock In \emph{2025 IEEE/CVF Conference on Computer Vision and Pattern Recognition (CVPR)}, pages 27968--27978. IEEE, 2025.

\end{thebibliography}
}
\clearpage
\section*{Overview}
This appendix presents comprehensive experimental details, evaluation details, and more visualization results. The content is organized into five main sections:

\begin{itemize}
\item Sec.~\ref{app:spatial_compression} presents the detailed proof of Proposition 3.1.

\item Sec.~\ref{LMTV} presents the details about generating key frames of large motion.

\item Sec.~\ref{seca} presents the training and inference details of our three diffusion models, LMTV-DM, LOF-DM, and MotionControlNet.

\item Sec.~\ref{secb} presents the evaluation details of quantitative results and human study.

\item Sec.~\ref{secc} visualizes more examples of generated intermediate frames via LumosFlow.

\item Sec.~\ref{secd} visualizes more examples of generated optical flows via LumosFlow.

\item Sec.~\ref{sece} visualizes more examples of generated videos via LumosFlow.

\item Sec.~\ref{sech} presents the quantitative results of OF-VAE and LOF-DM on different compression factors.

\end{itemize}

\section{Additional Analysis of Spatial Compression}
\label{app:spatial_compression}

\subsection{Spatial Variation and Reference Reconstruction}

We provide the detailed definitions and proof of Prop.3.1. Let $S\in\mathbb{R}^{K\times C\times H\times W}$ be a spatiotemporal signal. Its spatial mean is computed independently for each frame and channel:
\begin{equation}
\mu_{k,c} =\frac{1}{HW}\sum_{i=1}^{H}\sum_{j=1}^{W}S_{k,c,i,j}.
\end{equation}
We broadcast $\mu_{k,c}$ over the spatial dimensions when evaluating $S-\mu$.

The spatial differences are defined as
\begin{equation}
\begin{aligned}
(\nabla_x S)_{k,c,i,j} &=S_{k,c,i,j+1}-S_{k,c,i,j},\\
(\nabla_y S)_{k,c,i,j} &=S_{k,c,i+1,j}-S_{k,c,i,j}.
\end{aligned}
\end{equation}
Horizontal differences are evaluated for $j<W$, and vertical differences for $i<H$. No padding, temporal differences, or cross-channel differences are included. Accordingly,
\begin{equation}
\|\nabla_{xy}S\|_F^2 = \|\nabla_x S\|_F^2+\|\nabla_y S\|_F^2.
\end{equation}

For an integer $h\ge2$ dividing both $H$ and $W$, we partition each spatial plane into non-overlapping $h\times h$ blocks. For a block $B$, the encoder computes
\begin{equation}
\overline S_{k,c,B} = \frac{1}{h^2} \sum_{(i,j)\in B}S_{k,c,i,j}.
\end{equation}
The decoder reconstructs every position in that block using the same mean:
\begin{equation}
(\widehat S_h)_{k,c,i,j} = \overline S_{k,c,B}, \qquad (i,j)\in B.
\end{equation}
The encoder can be implemented by a depthwise convolution with kernel size and stride $h$ and weights $1/h^2$. The decoder performs nearest-neighbor upsampling. When $h=2^L$, the same construction can be implemented by $L$ successive $2\times2$ averaging convolutions with stride $2$, followed by corresponding upsampling stages.

We define the normalized reconstruction error as
\begin{equation}
D_S(h) = \frac{\|S-\widehat S_h\|_F^2}
{\|S-\mu\|_F^2},
\label{eq:app_normalized_error}
\end{equation}
assuming $\|S-\mu\|_F^2>0$. If this denominator is zero, every frame-channel plane is spatially constant and the reference construction reconstructs it exactly.

\subsection{Proof of Prop.3.1}
We first establish a one-dimensional estimate. Consider $x_1,\ldots,x_h$ with mean $\overline x=h^{-1}\sum_i x_i$. The pairwise variance identity gives
\begin{equation}
\sum_{i=1}^{h}(x_i-\overline x)^2 = \frac{1}{h}\sum_{i<j}(x_i-x_j)^2.
\label{eq:app_pairwise_variance}
\end{equation}
By telescoping and the Cauchy--Schwarz inequality,
\begin{equation}
\begin{aligned}
(x_i-x_j)^2 &\le
(j-i)\sum_{r=i}^{j-1}(x_{r+1}-x_r)^2\\
&\le h\sum_{r=i}^{j-1}(x_{r+1}-x_r)^2.
\end{aligned}
\end{equation}
Each adjacent difference at position $r$ appears in $r(h-r)$ pairs. Substituting into Eq.~\eqref{eq:app_pairwise_variance} therefore yields
\begin{equation}
\begin{aligned}
\sum_{i=1}^{h}(x_i-\overline x)^2 &\le \sum_{r=1}^{h-1} r(h-r)(x_{r+1}-x_r)^2\\
&\le \frac{h^2}{4} \sum_{r=1}^{h-1}(x_{r+1}-x_r)^2.
\end{aligned}
\label{eq:app_1d_bound}
\end{equation}

We next consider an $h\times h$ block $b$. Let $r_i=h^{-1}\sum_j b_{i,j}$ denote its row means and $\overline b=h^{-1}\sum_i r_i$ its block mean. The variance decomposes as
\begin{equation}
\begin{aligned}
\sum_{i,j}(b_{i,j}-\overline b)^2 ={}& \sum_{i,j}(b_{i,j}-r_i)^2\\
&+ h\sum_i(r_i-\overline b)^2.
\end{aligned}
\label{eq:app_block_decomposition}
\end{equation}
Applying Eq.~\eqref{eq:app_1d_bound} within each row gives
\begin{equation}
\sum_{i,j}(b_{i,j}-r_i)^2 \le \frac{h^2}{4} \sum_i\sum_{j=1}^{h-1} (b_{i,j+1}-b_{i,j})^2.
\end{equation}
Applying the same estimate to the row means, followed by Jensen's inequality, gives
\begin{equation}
\begin{aligned}
h\sum_i(r_i-\overline b)^2 &\le \frac{h^3}{4} \sum_{i=1}^{h-1}(r_{i+1}-r_i)^2\\
&\le \frac{h^2}{4} \sum_{i=1}^{h-1}\sum_j (b_{i+1,j}-b_{i,j})^2.
\end{aligned}
\end{equation}
Combining these estimates with Eq.~\eqref{eq:app_block_decomposition}, we obtain
\begin{equation}
\|b-\overline b\|_F^2 \le \frac{h^2}{4}\|\nabla_{\mathrm{int}}b\|_F^2, \label{eq:app_block_bound}
\end{equation}
where $\nabla_{\mathrm{int}}$ contains only adjacent differences within the block. Finally, summing Eq.~\eqref{eq:app_block_bound} over all blocks, frames, and channels yields
\begin{equation}
\begin{aligned}
\|S-\widehat S_h\|_F^2 &\le \frac{h^2}{4} \sum_{k,c,B} \|\nabla_{\mathrm{int}}S_{k,c,B}\|_F^2\\
&\le \frac{h^2}{4}\|\nabla_{xy}S\|_F^2.
\end{aligned}
\end{equation}
The last inequality follows because the within-block adjacent pairs form a subset of all valid spatial adjacent pairs. Dividing both sides by $\|S-\mu\|_F^2$ gives
\begin{equation}
D_S(h) \le \frac{h^2}{4} \frac{\|\nabla_{xy}S\|_F^2}{\|S-\mu\|_F^2} = \frac{h^2}{4}\gamma(S),
\end{equation}
which proves Prop.3.1.

\subsection{Implications for Spatial Compression}
For a target relative error $\epsilon\in(0,1)$,
Prop.3.1 provides the sufficient condition
\begin{equation}
h \le 2\sqrt{\frac{\epsilon}{\gamma(S)}},
\label{eq:app_sufficient_factor}
\end{equation}
subject to the integer and divisibility requirements on $h$. Thus, a smaller normalized spatial variation permits a larger spatial downsampling factor under the same sufficient error guarantee. This formalizes the connection between spatial coherence and a coarser spatial representation.

The comparison in Fig.3 evaluates this structural condition for optical flow and RGB video. For both representations, the numerator and denominator of $\gamma(S)$ are summed over all frames and channels before taking their ratio. Optical flow is evaluated using its displacement components rather than a color visualization. A common nonzero rescaling of a signal does not change $\gamma(S)$.

The proposition concerns the reference convolutional construction defined above. Lower $\gamma(S)$ provides a more favorable sufficient bound, but does not by itself establish an ordering of the actual reconstruction errors of different signals. We therefore use this analysis to motivate the spatial compression design and evaluate the reconstruction quality of the learned OF-VAE separately. The analysis addresses spatial compression only.

\section{Large Motion Text-to-Video Diffusion Model}
\label{LMTV}
Previous text-to-video diffusion models are capable of generating continuous videos with numerous frames; however, they lack the ability to produce key frames with larger intervals. These key frames are essential, as they significantly influence both the motion range and the overall scene of the video. Therefore, it is particularly important to generate key frames that are not only distinct but also consistent in subject matter. Unfortunately, prior hierarchical generation strategies do not explicitly account for these factors, which limits their effectiveness in capturing coherent narratives and maintaining visual continuity. To enhance the ability to generate key frames with substantial motion, we establish an additional training set consisting of videos at a lower frame rate, which results in videos with larger intervals. By fine-tuning the current text-to-video generation model, we can sample a video (key frames) that exhibits higher motion in accordance with the provided prompt.
Formally, it has:
\begin{equation}
    v \sim p_{\theta}(v \mid P), 
\end{equation}
where $v$ and $P$ are the generated video (a set of key frames) and the given text prompt.

\section{Training and Inference Details} 
\label{seca}
\begin{figure}[h]
    \centering
    \includegraphics[width=1\linewidth]{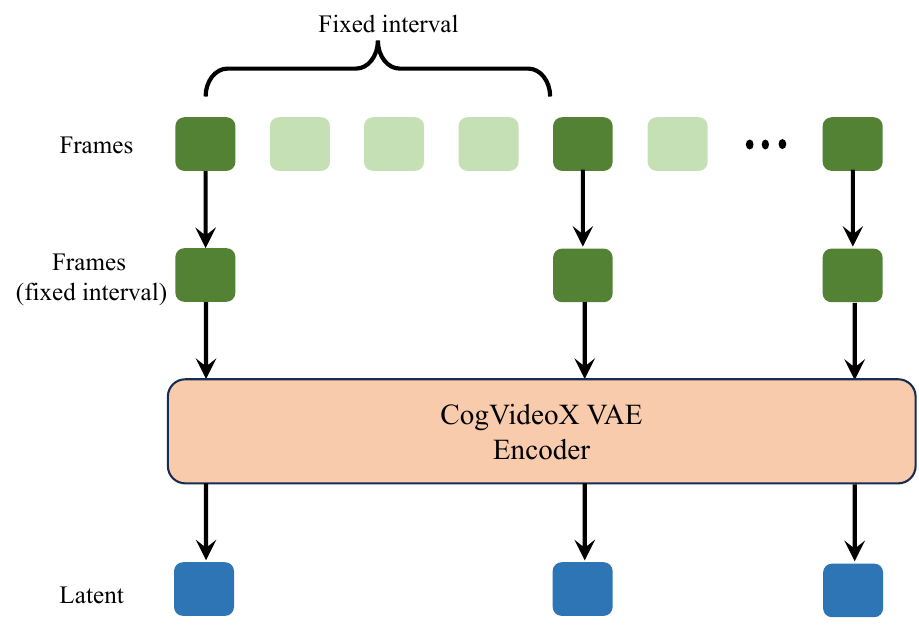}
    \caption{The overall encoding pipeline of the Large Motion Text-to-Video Diffusion Model. Given a set of frames, the selected frames (indicated in dark green) are sampled at fixed intervals. Subsequently, we employ the CogVideoX VAE to encode these frames in a frame-by-frame manner, resulting in the corresponding latent representations (shown in blue), which are then utilized in the conducted diffusion pipeline.}
    \label{stage1}
\end{figure}

\paragraph{Large Motion Text-to-Video Diffusion Model} Large Motion Text-to-Video Diffusion Model is fine-tuned based on the CogVideoX-5b Text-to-Video diffusion model. The training set, consisting of 600,000 samples, is filtered according to aesthetic scores and motion degrees, which are estimated using one-align and optical flow, respectively. During the fine-tuning phase, we randomly sample frames at fixed intervals and apply the CogVideoX video Variational Autoencoder (VAE) to encode these frames. In contrast to the original CogVideoX, which encodes four frames simultaneously, we encode the selected frames in a frame-by-frame manner, as larger intervals result in lower motion consistency. The overall encoding pipeline is depicted in Fig.~\ref{stage1}.

\paragraph{LOF-DM} The LOF-DM model is developed based on the DiT architecture, with hyperparameters detailed in Tab.~\ref{hyper_lofdm}. During the training phase, we employ a regularization strategy where we randomly drop semantic features and linear optical flow with a probability of 10\%. This technique aims to enhance the model's robustness by preventing overfitting. For the training process, we set the learning rate to $5\times10^{-5}$ and utilize a batch size of $128$, ensuring efficient model convergence. During the inference phase, we implement classifier-free guidance on the semantic features to improve the quality of the generated outputs.

\begin{table}[h]
    \centering
    \begin{tabular}{c|c}
    \toprule
    & LOF-DM\\
    \bottomrule
        Number of layers &  $30$\\
        Attention heads &  $32$\\
        Hidden size & $1920$\\
        Position encoding & sinusoidal\\
        Time Embedding Size & $256$\\
        Training Precision &Float32\\
    \bottomrule
    \end{tabular}
    \caption{Hyper-parameters of LOF-DM.}
    \label{hyper_lofdm}
\end{table}

\paragraph{MotionControlNet} 
The MotionControlNet is developed based on the CogVideoX-5b Image-to-Video diffusion model. In the training process, we configure the learning rate to $1\times10^{-4}$ and utilize a batch size of 8. Additionally, we randomly apply a negative prompt with a probability of 10\%. This strategy is implemented to enhance the robustness of the model and improve its overall performance in video generation tasks.

\section{Evaluation Details}
\label{secb}
For a fair comparison, we use the commonly used VBench~\cite{huang2023vbench} to evaluate the quality of generated videos as well as the commonly used metrics FVD, FID, and SSIM. VBench is designed specifically for benchmarking the performance of video generation models, providing a comprehensive suite of evaluation tools that facilitate the analysis of both qualitative and quantitative aspects of generated videos. In actual practice, we follow the instructions provided by VBench to divide the videos into distinct sets of frames and evaluate each set separately.

For the human evaluation, we randomly select 14 prompts and generate videos using various methods. We then invite 8 users to assess these videos based on four key metrics: Text Alignment, Frame Consistency, Dynamic Degree, and Video Quality. Text Alignment evaluates how accurately the videos correspond to the prompts, while Frame Consistency measures the stability of visual elements across frames. Dynamic Degree analyzes the level of motion captured in the videos, and Video Quality assesses overall visual appeal, including clarity and color fidelity. This comprehensive evaluation approach allows us to better understand the strengths and weaknesses of the generated videos, facilitating the improvement of our video generation methods.

\section{More Generated Intermediate Frames}
\label{secc}
\begin{figure*}
    \centering
    \includegraphics[width=1.\linewidth]{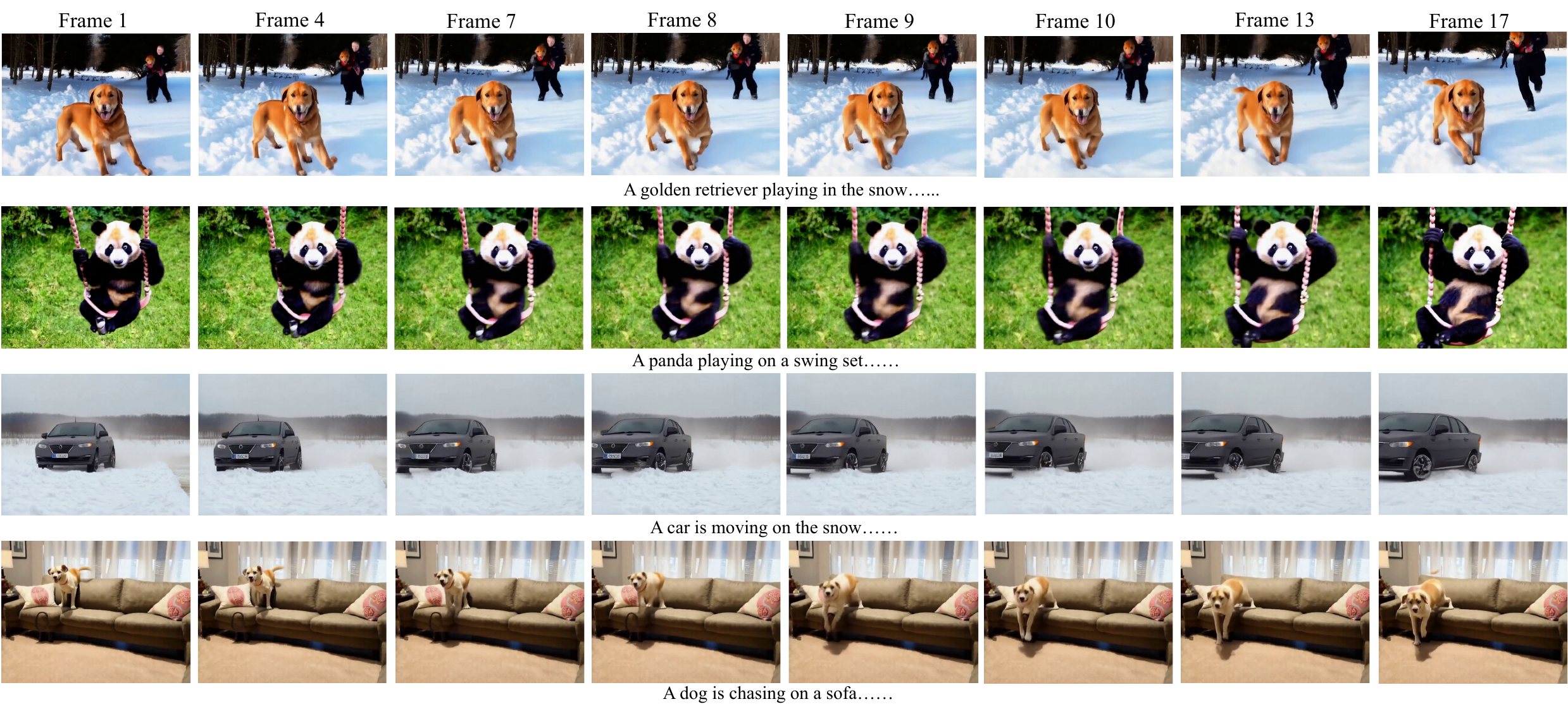}
    \caption{Generated intermediate frames via LumosFlow.}
    \label{supp_inter}
\end{figure*}

We present more generated intermediate frames of LumosFlow in Fig.~\ref{supp_inter}.

\section{More Generated Optical Flows}
\label{secd}
\begin{figure*}
    \centering
    \includegraphics[width=1.\linewidth]{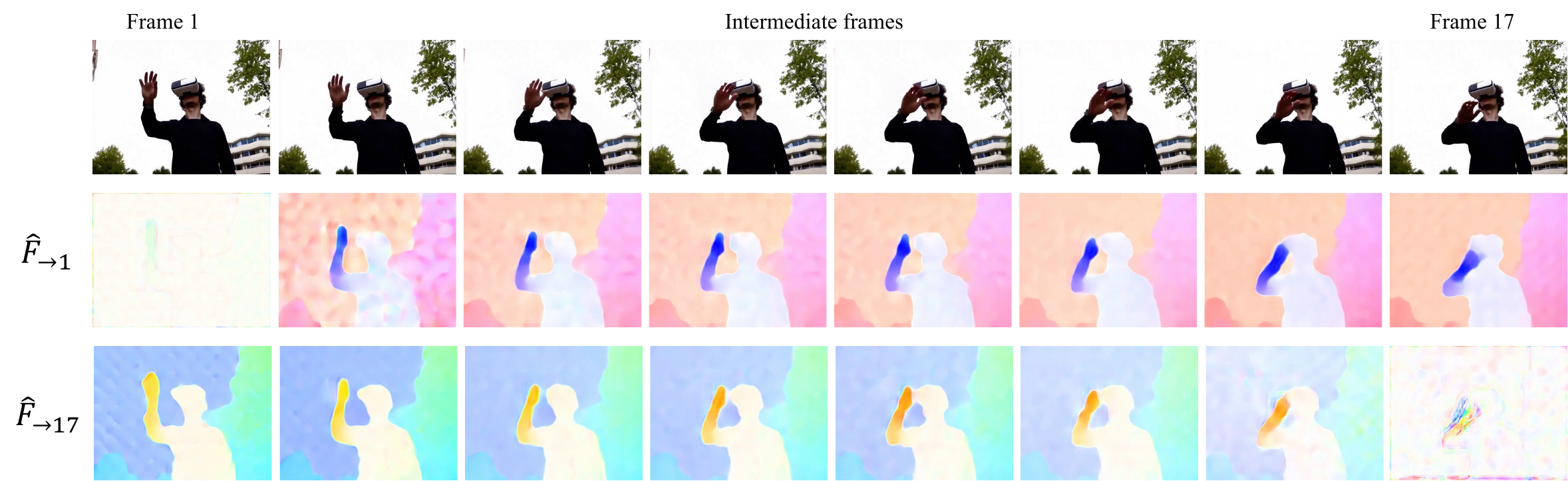}
    \caption{Generated optical flows and intermediate frames via the given first (Frame 1) and last frames (Frame 17). The first and last frames are generated by the LMTV-DM and the optical flows ($\hat{F}_{\rightarrow1}$ and $\hat{F}_{\rightarrow17}$) are generated by the LOF-DM. The intermediate frames are generated by the MotionControlNet.}
    \label{flow_gen}
\end{figure*}

We present more generated intermediate optical flows of LumosFlow in Fig.~\ref{flow_gen}.  Given the first and last frames, a rational movement involves the man's arm moving from the top left to the bottom right. The generated optical flow effectively captures and models this movement, demonstrating improved coherence and realism in the depiction of motion.

\section{More Generated Videos}
\label{sece}
\begin{figure*}
    \centering
    \includegraphics[width=.95\linewidth]{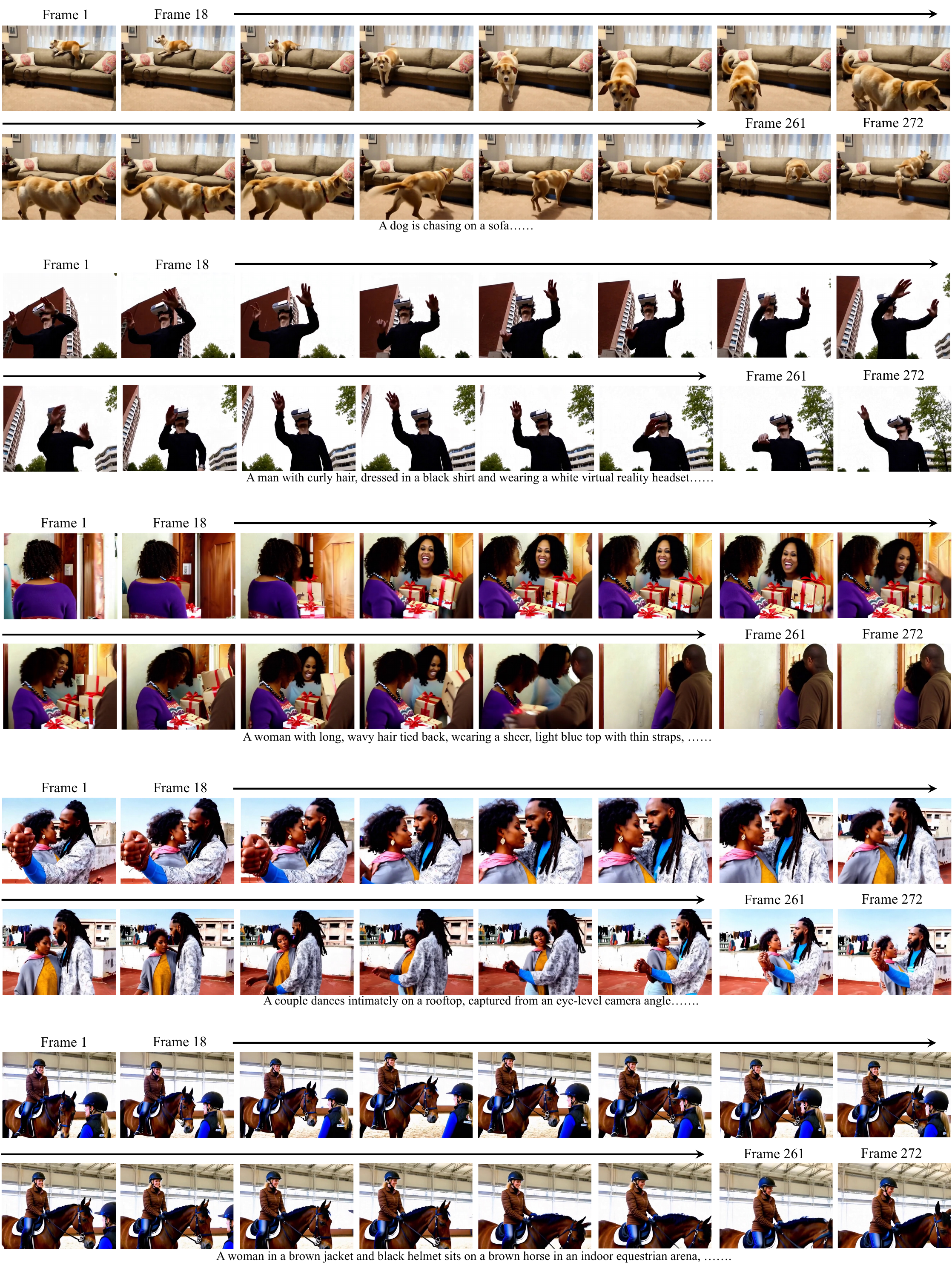}
    \caption{Generated videos via LumosFlow}
    \label{supp_gen}
\end{figure*}

We present more generated videos of LumosFlow in Fig.~\ref{supp_gen}. The generated videos exhibit a high degree of smoothness and feature complex movement dynamics.

\section{Compression performance}
\label{sech}
Tab.~\ref{tab:interpolation_epe} evaluates OF-VAE reconstruction and LOF-DM generation across interpolation factors from $4\times$ to $24\times$. OF-VAE reconstructs reference flows after compression, while LOF-DM generates flows conditioned on the first and last frames. Both reconstruction and generation EPE increase with the interpolation factor. Larger frame intervals can involve greater displacements and more complex motion, making accurate flow reconstruction and prediction more challenging. In particular, LOF-DM's forward and backward EPEs increase from $0.823$ and $0.844$ at $4\times$ to $4.698$ and $4.688$ at $24\times$, respectively, reflecting the increased difficulty of inferring intermediate motion across larger temporal gaps.

\begin{table*}[t]
    \centering
    \caption{Forward and backward EPE at different frame interpolation factors.}
    \label{tab:interpolation_epe}
    \small
    \setlength{\tabcolsep}{3pt}
    \begin{tabular}{l|cc|cc|cc|cc|cc|cc}
        \hline
        Method
        & \multicolumn{2}{c|}{$\times 4$}
        & \multicolumn{2}{c|}{$\times 8$}
        & \multicolumn{2}{c|}{$\times 12$}
        & \multicolumn{2}{c|}{$\times 16$}
        & \multicolumn{2}{c|}{$\times 20$}
        & \multicolumn{2}{c}{$\times 24$} \\
        \cline{2-13}
        & EPE$_{\mathrm{fwd}}$ & EPE$_{\mathrm{bwd}}$
        & EPE$_{\mathrm{fwd}}$ & EPE$_{\mathrm{bwd}}$
        & EPE$_{\mathrm{fwd}}$ & EPE$_{\mathrm{bwd}}$
        & EPE$_{\mathrm{fwd}}$ & EPE$_{\mathrm{bwd}}$
        & EPE$_{\mathrm{fwd}}$ & EPE$_{\mathrm{bwd}}$
        & EPE$_{\mathrm{fwd}}$ & EPE$_{\mathrm{bwd}}$ \\
        \hline
        OF-VAE
        & 0.439 & 0.456 & 0.520 & 0.592 & 0.978 & 0.983
        & 1.402 & 1.404& 1.424 & 1.543 & 1.955 & 1.972 \\
        LOF-DM
        & 0.823 & 0.844 & 1.306 & 1.243 & 1.523& 1.544
        & 2.872 & 2.899 & 2.943 & 2.954 & 4.698 & 4.688 \\
        \hline
    \end{tabular}
\end{table*}


\end{document}